\documentclass[review]{elsarticle}
\usepackage{lineno,hyperref}
\usepackage{ulem}
\usepackage{amsmath,amsfonts}
\usepackage{algorithmic}
\usepackage{algorithm}
\usepackage{array}
\usepackage{subfigure}

\usepackage{textcomp}
\usepackage{stfloats}
\usepackage{url}
\usepackage{verbatim}
\usepackage{graphicx}
\usepackage{booktabs}
\usepackage{color}
\journal{Knowledge-Based Systems}

\biboptions{authoryear}

\begin{document}

\begin{frontmatter}

\title{DPGNN: Dual-Perception Graph Neural Network for Representation Learning}
% \tnotetext[mytitlenote]{Fully documented templates are available in the elsarticle package on \href{http://www.ctan.org/tex-archive/macros/latex/contrib/elsarticle}{CTAN}.}

%\author{manuscript}
\author{Li Zhou}
\ead{li_zhou@std.uestc.edu.cn}
\author{Wenyu Chen\corref{mycorrespondingauthor}}
\cortext[mycorrespondingauthor]{Corresponding author}
\ead{cwy@uestc.edu.cn}
\author{Dingyi Zeng}
\ead{zengdingyi@std.uestc.edu.cn}
\author{Shaohuan Cheng}
\ead{shaohuancheng@std.uestc.edu.cn}
\author{Wanlong Liu}
\ead{liuwanlong@std.uestc.edu.cn}
\author{Malu Zhang}
\ead{maluzhang@uestc.edu.cn}
\author{Hong Qu}
\ead{hongqu@uestc.edu.cn}

\address{School of Computer Science and Engineering, University of Electronic Science and Technology of China, Chengdu, Sichuan 611731, P.R. China}
% \fntext[myfootnote]{Since 1880.}

% \author[mymainaddress,mysecondaryaddress]{Elsevier Inc}

% \author{Global Customer Service}

\begin{abstract}
Graph neural networks (GNNs) have drawn increasing attention in recent years and achieved remarkable performance in many graph-based tasks, especially in semi-supervised learning on graphs. However, most existing GNNs are based on the message-passing paradigm to iteratively aggregate neighborhood information in a single topology space. Despite their success, the expressive power of GNNs is limited by some drawbacks, such as inflexibility of message source expansion, negligence of node-level message output discrepancy, and restriction of single message space. To address these drawbacks, we present a novel message-passing paradigm, based on the properties of multi-step message source, node-specific message output, and multi-space message interaction. 
To verify its validity, we instantiate the new message-passing paradigm as a Dual-Perception Graph Neural Network (DPGNN), which applies a node-to-step attention mechanism to aggregate node-specific multi-step neighborhood information adaptively. Our proposed DPGNN can capture the structural neighborhood information and the feature-related information simultaneously for graph representation learning. Experimental results on six benchmark datasets with different topological structures demonstrate that our method outperforms the latest state-of-the-art models, which proves the superiority and versatility of our method.  To our knowledge, we are the first to consider node-specific message passing in the GNNs.
\end{abstract}

\end{frontmatter}

% \linenumbers
\section{Introduction}
There exists many graph data in life, such as knowledge graph \cite{li2022knowledge}, citation networks \cite{kipf2017semi, velivckovic2017graph} and traffic networks \cite{ta2022adaptive}, which traditional deep neural networks (DNNs) are very limited to process. And in recent years, Graph Neural Networks (GNNs) \cite{wu2020comprehensive} have been widely adopted in various graph-based tasks, such as node classification \cite{chen2022neighbor,yang2022semi}, graph classification \cite{gilmer2017neural, xie2020graph}, link prediction \cite{liu2021item, li2022collaborative, chen2022explainable}, clustering tasks \cite{kang2021structured, liao2022deep}, and knowledge tracing tasks \cite{SONG2021510, SONG2022108274}.

GNNs form an effective framework for the representation learning of graphs \cite{li2021deep, li2022transo}. And most existing GNNs \cite{kipf2017semi, velivckovic2017graph, wu2019simplifying} are mainly based on the message-passing paradigm, which iteratively aggregates neighborhood information to update a new representation of each node. However, one layer of GNNs only considers immediate neighbors and the performance degrades greatly when stacking multi-layer GNNs for a larger neighborhood receptive field. Recent studies have attributed this phenomenon to the over-smoothing problem \cite{li2018deeper, chen2020measuring}. To learn more effective node representations, various approaches dedicate to the breadth, depth, and strength of models. Some efforts focus on how to obtain multi-hop neighborhood information on a single-layer network \cite{abu2019mixhop, zhou2020weighted, yang2022semi}; some researches concentrate on designing deep GNN frameworks \cite{klicpera2018predict, liu2020towards, chen2020simple}; and some methods purpose to grow into a data augmenter \cite{rongdropedge, feng2020graph, verma2019graphmix}. 

However, despite GNNs revolutionizing graph representation learning, there are limitations to the expressive power of GNNs \cite{xu2018powerful,oono2019graph}. On the one hand, these models mostly follow the traditional message-passing paradigm in which the iterative operations are characteristic of the entire process. In this message-passing paradigm, the message source expansion is not flexible, and the node-level message output discrepancy for different neighborhood ranges is also not considered \cite{xu2018representation}. On the other hand, the existing message-passing paradigm is limited in a single topology space. Some experiments have verified that the original graph topology \cite{chen2020measuring} is the fundamental reason for the over-smoothing problem, because the nodes may receive messages with low information-to-noise ratio.

\begin{figure*}[t]
\centering
\includegraphics[width=1.0\textwidth]{figure/intro_graph_structure1.png} 
\caption{The topological views of six graph datasets. For intuitive presentation, we adopt different colors to distinguish nodes of different classes and set edges colorless. The number after the graph dataset name indicates the number of connected subgraphs of the graph. For example, there are 438 connected subgraphs in Citeseer, while Flickr is a big connected graph. Graph datasets Citeseer, ACM and CoraFull have similar topological structures, which are mainly composed of a large connected subgraphs (the middle part) and some small connected sub-graphs or isolated nodes (the surrounding part), while the topological structures of graph datasets UAI2010, Flickr and BlogCatalog are denser.}
\label{fig:graph_structure}
\end{figure*}

Most exiting GNNs primarily utilize graph topological structure for information propagation and representation learning, in which unfavourable topological structure may lead nodes receive too much noise after multiple steps of information propagation. Figure \ref{fig:graph_structure} shows topological views of six graph datasets  drawn by \textit{networkx} \cite{hagberg2008exploring}, in which nodes are positioned by Fruchterman-Reingold force-directed algorithm \cite{fruchterman1991graph}. Ideally, nodes of the same class desire to be more connected by edges, while there are many inter-class edges in some graph topologies. And some nodes exist in small connected subgraphs, including a multi-hop information capturing limitation.

In fact, graphs with nodes connected with different classes are common in the real world, and it is also common that nodes are strongly correlated but in different connected subgraphs. For example, different amino acid types are more likely to connect in protein structures \cite{zhu2020beyond}, fraudsters are more likely to connect to accomplices than to other fraudsters in online purchasing networks \cite{pandit2007netprobe}, and preferences for seasonal goods may be the same in different regions. To break the bottleneck caused by graph topological structures in GNNs, some works \cite{pei2020geom, wang2020gcn, wu2021dual} began to focus on other potential message-passing spaces to enrich graph representation learning and some methods \cite{franceschi2019learning, jiang2019semi, kang2021structured} propose graph structure learning techniques used for message passing. 
But most of these methods regard Graph Convolutional Networks (GCN) as the base encoder framework.

Motivated by observations like the above, in this paper, we summarize the drawbacks of the existing message-passing paradigm including inflexibility of message source expansion, negligence of node-level message output discrepancy, and restriction of single message space. Then we propose an improved message-passing paradigm that can support both node-specific multi-step message aggregation and multi-space interaction. Based on the new message-passing paradigm, we propose a novel Dual-Perception Graph Neural Network (DPGNN) for graph representation learning.

Our main contributions are summarized as follows:
\begin{itemize}
\item We formalize the existing message-passing paradigm, analyze its drawbacks, and present a novel improved message-passing paradigm based on the properties of multi-step message source, node-specific message output, and multi-space message interaction.
\item We instantiate the new message-passing paradigm as a Dual-Perception Graph Neural Network (DPGNN), which applies a node-to-step attention mechanism to aggregate node-specific multi-step neighborhood information adaptively and captures the structural neighborhood information and the feature-related information simultaneously for graph representation learning.
\item we apply DPGNN in the semi-supervised node classification task on six graph datasets with different topological structures. The experimental results demonstrate that our instantiated DPGNN outperforms related state-of-the-art GNNs, and we conduct analysis experiments to prove the superiority and versatility of our proposed message-passing paradigm.

\end{itemize}

The remainder of this article is organized as follows: the most related previous works are reviewed briefly in Section II; the improved message-passing paradigm is defined in Section III; the instantiated methods and experimental evaluations are presented in Sections IV and V respectively; and the conclusion is given in the last part.

\section{Related Works}
In this section, we review relevant research in our field. We divide most current GNNs into three categories: 1) GNNs based on graph topology, 2) GNNs based on node features, and 3) GNNs based on graph structure learning.

\subsection{GNNs Based on Graph Topology}
The original GNNs apply message passing mainly based on graph topology, which are improved on the breadth, depth, and strength of models for better performance. Inspired by the effectiveness of Convolutional Neural Networks (CNNs) on grid-like data such as images \cite{matsugu2003subject, he2016deep, karpathy2014large}, the vanilla GCN is proposed to show an efficient variant of convolutional neural networks which can operate directly on graphs. The propagation rule of vanilla GCN \cite{kipf2017semi} can be explained via the approximation of the spectral graph convolutions \cite{bruna2013spectral, defferrard2016convolutional}. GAT \cite{velivckovic2017graph} introduces an attention-based architecture to compute the hidden representations of each node in the graph. MixHop \cite{abu2019mixhop} can learn a general class of neighborhood mixing relationships by repeatedly mixing feature representations of neighbors at various distances. GraphMix \cite{verma2019graphmix} is a regularization method in which a fully-connected network is jointly trained with the graph neural network via parameter sharing and interpolation-based regularization. 

However, these methods are too shallow to consider high-order nodes and the size of the utilized neighborhood is hard to extend. To address this, Klicpera et al. \cite{klicpera2018predict} present a personalized propagation of neural predictions (PPNP) and its fast approximation APPNP based on personalized PageRank for a large neighborhood receptive field. Based on this, Chen et al. \cite{chen2020simple} propose a deep GCN model with initial residual and identity mapping. At each layer, initial residual constructs a skip connection from the input layer, while identity mapping adds an identity matrix to the weight matrix. And DAGNN \cite{liu2020towards} can adaptively incorporate information from large receptive fields by decoupling representation transformation and propagation. Feng et al. \cite{feng2020graph} first design a random propagation strategy to perform graph data augmentation and leverage consistency regularization in GRAND model to mitigate the issues of over-smoothing and non-robustness.

\subsection{GNNs Based on Node Features} 
Real-world graphs are noisy, i.e. adjacent nodes may not be similar, and similar nodes are not necessarily adjacent in a topological structure. Therefore, there are limitations in information aggregation and representation learning only considering topology space. In light of this, some works begin to focus on multi-space representation fusion \cite{xu2019adversarial, wu2021dual, CHANG2021106807}. In this work, we concentrate on the perspective of node features. Gao et al. \cite{gao2018large} propose a learnable graph convolutional layer (LGCL) to select a fixed number of neighboring nodes for each feature based on value ranking. In this way, the generic graph can be transformed into grid-like data that regular convolutional networks can operate on. So this approach considers feature importance to change the form of graph data, but the convolution aggregator in LGCL cannot be directly applied to graphs.
LA-GCN \cite{zhang2020feature} also concentrates on the importance of different features, introducing a learnable aggregator for GCN and proposing a new attention mechanism allowing both node-level and feature-level attention.
Sambaran et al. \cite{bandyopadhyay2020multilayered} propose an unsupervised algorithm MIRand, which creates a multi-layer graph (including structure layer and content layer) and employs a random walk that exploits the informativeness of a node by unifying its structure and attributes. In MIRand, the constructed content-layer graph is always directed and weighted.
RolEANE \cite{li2021deep} propose a neighbor optimization strategy, which is used to efficiently and seamlessly integrate the network topological structure and attribute information to improve representation learning performance.
Geom-GCN \cite{pei2020geom} maps node features to representation vectors, which can be considered as the position of each node in a latent continuous space. Then, based on the original graph and the latent space, Geom-GCN builds a structural neighborhood, and applies GCN to aggregate information of different neighborhoods. Both AMGCN \cite{wang2020gcn} and SCRL \cite{liu2021self} construct feature graphs by input features of nodes and then apply GCN encoders to extract effective information. 
The difference is that the former learns specific and common embeddings from both feature and topology graph, constraining their consistency and diversity during training. And the latter mainly designs a self-supervised loss to maximize the agreement of the embeddings of the same node in two view graphs.

\subsection{GNNs Based on Graph Structure Learning} 
In addition, several efforts have been made to alleviate the imperfection that GNNs rely on the good quality of raw graph topological structure. So some graph structure learning methods are proposed. For homogeneous graphs, Franceschi et al. propose a framework LDS \cite{franceschi2019learning} that can learn the graph structure and the parameters of a GNN simultaneously. GLCN \cite{jiang2019semi} aims to learn an optimal graph structure by integrating both graph learning and graph convolution in a unified network architecture. And IDGL \cite{chen2020iterative} is proposed to jointly learn the graph structure and graph embeddings by optimizing a joint loss combining both task prediction loss and graph regularization loss. For heterogeneous graphs, GTN \cite{zhao2021heterogeneous} is capable of generating new graph structures, which involves identifying useful connections between unconnected nodes on the original graph. HGSL \cite{yun2019graph} is designed to learn an optimal heterogeneous graph structure.

These methods focus on the reconstruction and consideration of graph neural network frameworks for better graph representation learning. However, their message-passing paradigm remains an iterative design and ignores the node-level message output discrepancy. In light of this, we further formalize and analyze the existing message-passing paradigm, and redefine an improved paradigm.

\section{An improved message-passing paradigm}

% \subsection{Preliminaries}
A graph is formally defined as $\mathcal{G}=(\mathcal{V},\mathcal{E})$, where $\mathcal{V}=\mathcal{V}_l\cup \mathcal{V}_u$ represents the union of $N_l$ labeled nodes ($\mathcal{V}_l$) and $N_u$ unlabeled nodes ($\mathcal{V}_u$), $\mathcal{E}\subseteq \mathcal{V}\times \mathcal{V}$ is a set of edges between nodes, $N=\vert \mathcal{V}\vert$ and $M=\vert \mathcal{E}\vert$ represent the number of nodes and edges respectively. The node features are denoted as $\mathbf{X}=\{\boldsymbol{x}_0,\;\boldsymbol{x}_1,\dots,\boldsymbol{x}_{N-1}\}\in \mathbb{R}^{N\times d}$, where $d$ is the dimension of node features and each node feature $\boldsymbol{x}_i$ is a non-negative vector. $\mathbf{A} \in \mathbb{R}^{N\times N}$ denotes the adjacency matrix of $\mathcal{G}$, with each element $\mathbf{A}_{ij}=1$ associating there exists an edge between node $i$ and node $j$, otherwise $\mathbf{A}_{ij}=0$. In an undirected graph, $\mathbf{A}_{ij}=\mathbf{A}_{ji}$. Noteworthy, $\mathbf{\hat{A}}=\mathbf{\tilde{D}}^{-1/2}\mathbf{\tilde{A}\tilde{D}}^{-1/2}$ is the symmetric normalization of the adjacency matrix, in which $\mathbf{\tilde{A}}=\mathbf{A}+\mathbf{I}_{N}$ indicates that all nodes in the graph have added self-loop edges, $\mathbf{\tilde{D}}_{ii}=\sum{_j\mathbf{\tilde{A}}_{ij}}$ is the diagonal degree matrix.  The labels of $N_l$ labeled nodes are denoted as $\mathbf{Y}_l=\{\boldsymbol{y}_0, \boldsymbol{y}_1, \dots, \boldsymbol{y}_{N_l-1}\}\in \mathbb{R}^{N_l \times C}$, where $\boldsymbol{y}_i$ is a one-hot vector and $C$ is the number of classes. 

For semi-supervised classification, only a few nodes observe their labels, while other nodes' labels are missing, i.e., $0<N_l\ll N_u$. The message-passing space mostly relies on the given $\mathcal{G}$. So the task is to design a graph neural network $\mathbf{Z}=f(\mathbf{X},\mathbf{A};\varTheta )$ to learn the node representations and predict the label of unlabeled nodes finally, where $\varTheta $ represents the trainable parameters.

The core of GNNs is the message-passing paradigm, which defines the way of information interaction between nodes and affects the representation learning ability of GNNs. In this section, we first discuss the existing message-passing paradigm of GNNs, analyze its limitations, and present an improved message-passing paradigm. For the convenience of subsequent discussion, we define some notations. 
\\
\textbf{Notation 1.} \textit{Let $d\left(i,j \right)$ be the shortest path length from node $i$ to node $j$ in the $\mathcal{G}$. Particularly, $d\left(i,i \right)=0$, $d\left(i,j \right)=1$ if $(i,j)\in \mathcal{E}$, and $d\left(i,j \right)=\infty$ if there is no path between node $i$ and node $j$.}
\\
\textbf{Notation 2.} \textit{Let $c_{\iota}\left( i,j \right)$ count the number of paths with length $\iota$ from node $i$ to node $j$. Particularly, for any $\iota$, $c_{\iota}\left( i,j \right)=0$ if $d\left(i,j \right)=\infty$.}
\\
\textbf{Notation 3.} \textit{Let $\mathcal{N} _{\iota}\left( i \right)$ be the set of nodes in which $\forall j\in \mathcal{N} _{\iota}\left( i \right)$ satisfies $d\left( i,j \right) =\iota$. Particularly, $\mathcal{N} _0\left( i \right) =\left\{ i \right\}$, and $\mathcal{N} _1\left( i \right)$ denotes the immediate neighbors of node $i$.}
\\
\textbf{Notation 4.} \textit{Let $\mathcal{J} _{\iota}\left( i \right)$ be the set of nodes in which $\forall j\in \mathcal{J} _{\iota}\left( i \right)$ satisfies $c_{\iota}\left( i,j \right) >0$. Particularly, $\mathcal{N} _{\iota}\left( i \right) \subseteq \mathcal{J} _{\iota}\left( i \right)$.}

\subsection{The existing message-passing paradigm of GNNs}
The goal of GNNs is to learn meaningful node representations, which can contain enough abundant but distinguishable information. And the message-passing paradigm of most existing GNNs is iterative passing and aggregation of local neighborhood messages based on topological structure. Let $\mathfrak{m} _{i}^{\left( t \right)}$ be the message of node $i$ obtained in the iteration $t$, and its message aggregation can be expressed simply as the sum of the immediate neighborhood messages:

\begin{equation}
    \mathfrak{m} _{i}^{\left( t \right)}=\alpha _i\mathfrak{m} _{i}^{\left( t-1 \right)}+\sum_{j\in \mathcal{N}_1 \left( i \right)}{\beta _j}\mathfrak{m} _{j}^{\left( t-1 \right)},
\label{eq:existing_mp}
\end{equation}
where $\alpha _i$ controls the amount of message retention at node $i$ and $\beta _j$ controls the amount of message input from node $j$. 

As shown in Figure \ref{fig:story} (a), the existing message-passing paradigm in each iteration, akin to a star-shaped pattern, can define the breadth of each message-passing iteration. The iteration of the existing message-passing paradigm enables message interaction between nodes that have reachable paths but are not directly connected. We regard this phenomenon as message transitivity. And as shown in Figure \ref{fig:story} (b), message transitivity, akin to a chain-shaped pattern, can ensure the depth of message passing over multiple iterations.

\begin{figure*}[t]
\centering
\includegraphics[width=1\textwidth]{figure/story.png} 
\caption{(a) The existing message-passing paradigm, akin to a star-shaped pattern, defines the breadth of each message-passing iteration. (b) Message transitivity, akin to a chain-shaped pattern, ensures the depth of message passing over multiple iterations. (c) The boosting breadth of incoming messages requires increasing the depth of message passing. (d) The message-passing paradigm is analyzed from three aspects: message retention, message input, and message output. (e) An example illustrating the heterogeneity of different message-passing spaces.}
\label{fig:story}
\end{figure*}

\subsection{Drawbacks of the existing message-passing paradigm} 
Although most GNNs follow the message-passing paradigm mentioned above, there still exists some drawbacks, which are summarized in three aspects.

(1) \textbf{Inflexibility of message source expansion:} In each iteration, each node $i$ can only aggregate its immediate neighborhood information, and its aggregation paradigm is static, i.e. the amount of message retention $\alpha_i$ and the amount of message input $\beta_j$ for node $j$ are independent and unchanged. Therefore, on the one hand, boosting the breadth of incoming messages requires increasing the depth of message passing. As shown in Figure \ref{fig:story} (c), when breadth is realized by increasing depth, the incoming messages increase exponentially, in which each node tends to obtain the entire graph information, and becomes indistinguishable finally. On the other hand, although iteratively transitivity can introduce messages from far-hop nodes, it also weakens messages from near-hop nodes.

(2) \textbf{Negligence of node-level message output discrepancy:} From Eq.\ref{eq:existing_mp}, we can observe that the message-passing paradigm only focuses on message retention and input in each iteration, ignoring the amount of message output for each node after message aggregation. As shown in Figure \ref{fig:story} (d)\footnote{All the message-passing paradigm can be generalized as $\mathbf{m} _{i}^{\left( t \right)}=o_i(\alpha _i\mathbf{m} _{i}^{\left( t-1 \right)}+\sum_{j\in \mathcal{N} _1\left( i \right)}{\beta _j}\mathbf{m} _{j}^{\left( t-1 \right)})$. When $o_i=1$, the above equation is simplified to the existing message-passing paradigm (i.e., Eq.\ref{eq:existing_mp}).},  in the existing message-passing paradigm, the output control coefficient $o_i$ of the aggregated message sum in each iteration for each node $i$ is always set to $1$, while in fact, each node, may pay different attention to different neighborhood ranges. See \ref{ap:1} for the further instance analysis.

(3) \textbf{Restriction of single message space.} In most GNNs, message passing is based on the original topological structure. However, the distance between nodes may be varied in different message spaces. For example, in the topology space, short-distance dependent nodes are not necessarily feature-correlated with each other, while long-distance dependent nodes may be very close in the feature space. More specifically, as shown in the Figure \ref{fig:story} (e), in topology space, the paper GAT \cite{velivckovic2017graph} cites Transformer \cite{vaswani2017attention} and GRAND \cite{feng2020graph} cites GAT, but the correlation between GRAND and Transformer is very weak, and the information interaction between them may introduce noise information to each other. However, in feature space, the content of AM-GCN \cite{wang2020gcn} is similar to GAT, and GAT is similar to GRAND, so AM-GCN is likely to be related to GRAND. Therefore, with the heterogeneity of different spaces, messages passing in only one space will lose other latent information interaction opportunities.

\subsection{The proposed message-passing paradigm}
To resolve the aforementioned drawbacks, we extend the original message-passing paradigm from three aspects: multi-step message source, node-specific message output, and multi-space message interaction. 

\textbf{Multi-step message source:} we broaden the message sources setting without extra iteration, in which the incoming messages are not limited to the immediate neighborhood. We regard the nodes which can be reached at the given step length $\iota$ as the message sources. To enrich message interactions from nodes of different step lengths, we define $\iota$ as the range of 0 to $L$. When $\iota=0$, the incoming messages come from the node itself. The aggregated message of node $i$ is formally defined as follows:
\begin{equation}
    \bar{\mathfrak{m}} _i=\sum_{\iota =0}^L{\left( \sum_{j\in \mathcal{J} _{\iota}\left( i \right)}{w_{\iota ,ij}}\mathfrak{m} _j \right)},
\label{eq:bread_of_message}
\end{equation}
where $0<w_{\iota, ij}<1$ is a transfer coefficient for evaluating the amount of message transfer from node $j$ to node $i$ under step length $\iota$. We use '-' to distinguish between messages before aggregation and messages after aggregation.

\textbf{Node-specific message output:} we focus on the differentiated control of node-specific message aggregation under various step lengths. So we further extend the message-passing paradigm (Eq.\ref{eq:bread_of_message}) as follows:

\begin{equation}
\bar{\mathfrak{m}} _i=\sum_{\iota =0}^L{o_{i\iota}\left( \sum_{j\in \mathcal{J} _{\iota}\left( i \right)}{w_{\iota ,ij}}\mathfrak{m} _j \right)},
\label{eq:control_of_output}
\end{equation}
where $\sum_{\iota =0}^L{o_{i\iota}=1}$ and $o_{i\iota}>0$, indicating that node $i$ pays attention to the message sources in step $\iota$.
% $o_{v\iota}$ , and $\sum_{\iota =0}^L{o_{v\iota}=1}$ is used to normalize the final message output.

\textbf{Multi-space message interaction:} we expand the original message-passing paradigm from a single space to multiple spaces, which can improve the correlation degree of node messages. The formalization is defined as:
\begin{equation}
\mathfrak{m} _i=Agg\left( \left\{ \bar{\mathfrak{m}} _{i}^{\varsigma}| \varsigma \in \varOmega \right\} \right),
\label{eq:message_space}
\end{equation}
where $Agg(\cdot)$ is a message aggregator function, $\varOmega$ is the message-passing space set, $\bar{\mathfrak{m}} _{i}^{\varsigma}$ denotes the aggregated message of node $i$ in corresponding space $\varsigma$. In this paper, we realize $\varOmega =\left\{ t,f \right\}$, related to the topology space and feature space. So $\bar{\mathfrak{m}} _i^{t}$ is derived from message aggregation in the original topology graph $\mathcal{G}^t$, while $\bar{\mathfrak{m}} _i^{f}$ comes from message aggregation in a new feature graph $\mathcal{G}^f$. The new feature graph is constructed by the similarity of node features with KNN algorithm \cite{wang2020gcn,liu2021self}.

\textbf{Overall}, we summarize and propose a new generalized message-passing paradigm, defined as follows:

\begin{equation}
    \begin{array}{c}
        \mathfrak{m} _i=Agg\left( \left\{ \bar{\mathfrak{m}} _{i}^{\varsigma}| \varsigma \in \varOmega \right\} \right), \\
         \bar{\mathfrak{m}}_{i}^{\varsigma}=\sum_{\iota =0}^{L^{\varsigma}}{o_{i\iota}^{\varsigma}\left( \sum_{j=0}^{N-1}{w_{\iota, ij}^{\varsigma}\mathfrak{m} _{j}^{\varsigma}} \right)}.
    \end{array}
    \label{eq:new_paradigm}
\end{equation}

% \begin{equation}
%      \mathfrak{m} _i=Agg\left( \left\{ \bar{\mathfrak{m}} _{i}^{\varsigma}| \varsigma \in \varOmega \right\} \right),
% \label{eq:new_paradigm1}
% \end{equation}
    
% \begin{equation}
%      \bar{\mathfrak{m}}_{i}^{\varsigma}=\sum_{\iota =0}^{L^{\varsigma}}{o_{i\iota}^{\varsigma}\left( \sum_{j=0}^{N-1}{w_{\iota, ij}^{\varsigma}\mathfrak{m} _{j}^{\varsigma}} \right)}.
% \label{eq:new_paradigm2}
% \end{equation}
Particularly, a carefully designed $w_{\iota, ij}^{\varsigma}$ should exhibit the following properties:

(1) \textbf{Message closeness:} $w_{\iota, ij_1}^{\varsigma}\leq w_{\iota, ij_2}^{\varsigma}$ if $d\left( i,j_1 \right) \geq d\left( i,j_2 \right)$ in the message space $\varsigma$, which implies that the closer the distance between nodes, the more their information interaction. 

(2) \textbf{Message denseness:} $w_{\iota, ij_1}^{\varsigma}\leq w_{\iota, ij_2}^{\varsigma}$ if $c_{\iota}\left( i,j_1 \right) \leq c_{\iota}\left( i,j_2 \right)$ in the message space $\varsigma$, which implies that the more accessible paths between nodes, the greater their potential association.

(3) \textbf{Message irrelevance:} $w_{\iota, ij}^{\varsigma}=0$ if $c_{\iota}\left( i,j \right) =0$,  which implies that if there is no reachable path of length $\iota$ between node $i$ and node $j$, there will be no message passing.

In the new message-passing paradigm, each node $i$ can acquire abundant and unique information. Specifically, on the one hand, the new message-passing paradigm directly increases the message receptive field without iteration operations and provides a diversity of message sources. On the other hand, each node $i$ can capture rich information from different message-passing space.

\section{Dual-Perception Graph Neural Network}
Generally, there are many different ways of implementing our improved message-passing paradigm, leading to GNNs with different expressive powers. In this section, we propose a novel Dual-Perception Graph Neural Network (DPGNN) that is an instantiation of our improved message-passing paradigm.

To facilitate implementation, we reconstruct Eq.\ref{eq:new_paradigm} as follows:

\begin{equation}
\bar{\mathfrak{m}}_{i}^{\varsigma}=\sum_{\iota =0}^{L^{\varsigma}}{\sum_{j=0}^{N-1}{\begin{array}{c}
	o_{i\iota}^{\varsigma}w_{\iota ,ij}^{\varsigma}\mathfrak{m} _{j}^{\varsigma}\\
\end{array}}}=\sum_{j=0}^{N-1}{\mathbf{M}_{ij}^{\varsigma}\mathfrak{m} _{j}^{\varsigma}},
\end{equation}

\begin{equation}
\mathbf{M}_{ij}^{\varsigma}=\left\{
\begin{aligned}
\sum_{\iota =0}^{L^{\varsigma}}{o_{i\iota}^{\varsigma}w_{\iota,ij}^{\varsigma}} & , & if\;{d\left( i,j \right) \ne \infty}, \\
0 & , & otherwise,
\end{aligned}
\right.
\end{equation}
where $\mathbf{M}_{ij}^{\varsigma}$ quantifies the message interaction between node $i$ and $j$ in the message-passing space $\varsigma$, and $\mathbf{M}^{\varsigma}=\left(\mathbf{M}_{ij}^{\varsigma} \right) _{i,j\in \mathcal{V}}$ is a soft-weighted adjacency matrix.

\begin{figure*}[htb]
\centering
\includegraphics[width=0.9\textwidth]{figure/model_framework20220407.png} 
\caption{The framework of DPGNN model. 1) \textit{KNN Module}: which constructs a feature graph $\mathcal{G}^f$ by the similarity of node features\cite{wang2020gcn,liu2021self}. 2) \textit{Soft-Weighted Adjacency Matrix Learning Module}: which learns two soft-weighted adjacency matrices $\overline{\mathbf{M}}^t$ and $\overline{\mathbf{M}}^f$ based on the two message-passing space. 3) \textit{Dual-Perception Representation Learning Module}: which 
captures the structural neighborhood information and the feature-related information simultaneously for each node. 4) \textit{Optimization Module}: which defines the loss function for DPGNN.}
\label{fig:model_framework}
\end{figure*}

Therefore, the key of DPGNN is: 1) how to obtain $\mathbf{M}^{\varsigma}$ in each message-passing space $\varsigma$, and 2) how to learn the final representation of nodes. This corresponds exactly to its two modules: \textit{Soft-Weighted Adjacency Matrix Learning} $\&$ \textit{Dual-Perception Representation Learning}. The framework of DPGNN is shown in Figure \ref{fig:model_framework}.

\begin{figure*}[htb]
\centering
\includegraphics[width=0.45\linewidth]{figure/MHGG.png} 
\caption{Soft-Weighted Adjacency Matrix Learning Module}
\label{fig:MHGG}
\end{figure*}

\subsection{Soft-Weighted Adjacency Matrix Learning}

In this part, we will introduce the soft-weighted adjacency matrix learning method $\varphi :\mathcal{G} \left( \mathcal{V} ,\mathcal{E} \right) \rightarrow \mathbf{M}$ for a general message-passing space $\mathcal{G}$. We firstly provide a definition of $w_{\iota, vu}$ that satisfies the properties of message closeness, message denseness, and message irrelevance. We find that $w_{\iota ,ij}\propto \mathbf{A}_{ij}^{\iota}$, where $\mathbf{A}^{\iota}$ is a matrix product of $\iota$ copies of $\mathbf{A}$, and $\mathbf{A}_{ij}^{\iota}=c_{\iota}\left( i,j \right)$. To normalize message aggregation, we apply $\mathbf{\hat{A}}$ as the transfer coefficient base, so we define $w_{\iota, ij}$ as follows:
\begin{equation}
    w_{\iota ,ij}=\mathbf{\hat{A}}_{ij}^{\iota}.
\end{equation}
Directly, as shown in Figure \ref{fig:MHGG}, we obtain the matrix set $\mathbb{A} =\left\{ \mathbf{\hat{A}}^0, \mathbf{\hat{A}}^1, …, \mathbf{\hat{A}}^L \right\} $, which involves a series of neighborhood information of different step lengths. 
% To acquire node-specific multi-hop information propagation, we design a node-to-hop attention mechanism, which is defined as:
Then we compute the convex combination of each node by 1×1 convolution with non-negative weights by using node-to-step attention mechanism: 
\begin{equation}
    \mathbf{M}_{ij}=\sum_{\iota =0}^{L}{o_{i\iota}\mathbf{\hat{A}}_{ij}^{\iota}},
\label{eq:soft-weighted}
\end{equation}
where $o_{i\iota}$ is the element of the learnable $\mathbf{W}_o$ after softmax operation, $\mathbf{W}_o\in \mathbb{R}^{N\times L}$, and $o_{i\iota}$ can represents the attention weight of node $i$ to the information in step length $\iota$.

Intuitively, we can regard $\mathbf{M}$ as the weighted adjacency matrix of a generated graph structure, which expands the original graph structure by adding edges to nodes that are not directly connected but have an accessible path, acting like a message-passing extender and controller. 
However, the added edges cause the graph to become denser, which behaves like a double-edged sword. To avoid redundancy of information aggregation, we introduce a random graph sparse strategy. Formally, we first randomly sample a binary mask $\epsilon _{
ij}\sim Bernoulli\left( 1-p \right)$ for each node pair. Second, we obtain the sparse weighted adjacency matrix $\overline{\mathbf{M}}$ by multiplying each edge weight with its corresponding mask:
\begin{equation}
    \overline{\mathbf{M}}_{ij}=\epsilon _{ij}\cdot \mathbf{M}_{ij}.
\end{equation}
Finally, we scale $\overline{\mathbf{M}}$ with the factor of $\frac{1}{1-p}$ to guarantee the weighted adjacency matrix is in expectation equal to $\mathbf{M}$.

By this strategy, we increase the randomness and the diversity of message passing, avoid the strong dependence of node representation learning on the newly generated graph structure, and reduce the computational complexity of message aggregation. Note that the random graph sparse strategy is only performed during training. During inference, we directly set $\overline{\mathbf{M}}$ as the original $\mathbf{M}$. In this way, based on topology graph and feature graph, we construct two soft-weighted adjacency matrix $\overline{\mathbf{M}}^t$ and $\overline{\mathbf{M}}^f$.

\begin{figure*}[t]
\centering
\includegraphics[width=0.75\linewidth]{figure/DPGNN.png}
\caption{Dual-Perception Representation Learning Module}
\label{fig:DPGNN}
\end{figure*}

\subsection{Dual-Perception Representation Learning}
In this part, we propose a Dual-Perception representation learning method based on topology space and feature space simultaneously, which adapt to the new message-passing paradigm. As shown in Figure \ref{fig:DPGNN}, we firstly apply a weight-shared layer to transform each node feature into a low-dimensional space:
\begin{equation}
\boldsymbol{h}_i=\mathrm{ReLU}\left(\boldsymbol{x}_i\mathbf{W}_h+\boldsymbol{b}_h \right),
\label{eq:weight-shared}
\end{equation}
where $\mathbf{W}_h\in \mathbb{R}^{d\times d_h}$, $\boldsymbol{b}_h\in \mathbb{R}^{d_h}$ are learnable parameters, $d$ is the dimension of node features, and $d_h$ is the hidden dimension. Secondly, two weight-exclusive layers are adopted to learn the node representations based on topology space and feature space in parallel:

\begin{equation}
    \boldsymbol{m}_i^t=\boldsymbol{h}_i \mathbf{W}_t+\boldsymbol{b}_t,
\end{equation}

\begin{equation}
    \boldsymbol{m}_i^f=\boldsymbol{h}_i \mathbf{W}_f+\boldsymbol{b}_f,
\end{equation}
where $\mathbf{W}_t,\mathbf{W}_f\in \mathbb{R}^{d_h\times C}$, $\boldsymbol{b}_t,\boldsymbol{b}_f \in \mathbb{R}^{C}$, $C$ represents the number of classes. Now $\boldsymbol{m}_i^t$ and $\boldsymbol{m}_i^f$ represent the message of node $i$ in corresponding space. Then we apply information aggregation for each message-passing space:

\begin{equation}
    \boldsymbol{\bar{m}}_{i}^t=\sum_{j=0}^{N-1}{\overline{\mathbf{M}}_{ij}^t\boldsymbol{m}_{j}^t},
\end{equation}

\begin{equation}
    \boldsymbol{\bar{m}}_{i}^f=\sum_{j=0}^{N-1}{\overline{\mathbf{M}}_{ij}^f\boldsymbol{m}_{j}^f}.
\end{equation}
Finally, we choose a message aggregator to obtain the final representation $\boldsymbol{z}_{i}$ of node $i$:
\begin{equation}
    \boldsymbol{z}_i=Agg(\boldsymbol{\bar{m}}_{i}^t, \boldsymbol{\bar{m}}_{i}^f) \in \mathbb{R}^{C}.
\label{eq:final_representation}
\end{equation}
The aggregator function mainly integrates topology-based node representations and feature-based node representations. In practice, we primarily examined three aggregator functions: Attention aggregator, Mean-pooling aggregator and Max-pooling aggregator. Based on DPGNN, node representations learning can capture a large range of neighborhood information in two different spaces adaptively.

% Based on this module, node representations learning can capture information in two different spaces.

\subsection{Optimization}

To verify the ability of representation learning, we apply DPGNN in the semi-supervised node classification task. Inspired by recent advances in semi-supervised learning \cite{berthelot2019mixmatch, feng2020graph}, we define the loss function as two parts: Cross-Entropy loss and Low-Entropy loss, which helps to train a low-entropy and strong-robustness model.

\textbf{Cross-Entropy Loss:} This part follows the loss function of most semi-supervised learning node classification tasks, only focusing on $n_l$ labeled nodes used for training. Firstly, the predicted probability $\boldsymbol{p}_i$ of each node $i$ is obtained from the final node representation:

\begin{equation}
    % \boldsymbol{p}_{vc}=\frac{\exp \left( \boldsymbol{z}_{vc} \right)}{\sum_{c=0}^{C-1}{\exp \left( \boldsymbol{z}_{vc} \right)}},
    \boldsymbol{p}_i=softmax \left( \boldsymbol{z}_i \right) =\exp \left( \boldsymbol{z}_{i}^{c} \right) /\sum_{c=0}^{C-1}{\exp \left( \boldsymbol{z}_{i}^{c} \right)},
\label{eq:probability}
\end{equation}
where $\boldsymbol{z}_i^c$ is $c$-th element of $\boldsymbol{z}_i$. Then the Cross-Entropy loss is defined as follows:
\begin{equation}
    CE\left( \mathbf{P}_{l},\mathbf{Y}_{l} \right) =-\sum_{i=0}^{N_l-1}{\boldsymbol{y}_{i}^{\mathrm{T}}\log \left( \boldsymbol{p}_i \right)},
\label{eq:CE}
\end{equation}
where $\mathbf{P}_{l}=\{\boldsymbol{p}_0, \boldsymbol{p}_1, \dots, \boldsymbol{p}_{N_l-1}\}\in \mathbb{R}^{N_l\times C}$ and $\mathbf{Y}_{l}=\{\boldsymbol{y}_0, \boldsymbol{y}_1, \dots, \boldsymbol{y}_{N_l-1}\}\in \mathbb{R}^{N_l\times C}$ are the predicted probability distribution and true labels of $N_l$ labeled nodes respectively.

\textbf{Low-Entropy Loss:} As a model for classification, we expect to realize an ideal model that is very certain about the predicted results. This means that the predicted probability distribution tends to have a low entropy. So we apply a sharpening function $sharpen\left( \cdot \right)$ to reduce the entropy of the label distribution \cite{berthelot2019mixmatch} to obtain an anchor label $\boldsymbol{\tilde{p}}_i$, the element of which is defined as follow:

\begin{equation}
    % \boldsymbol{\tilde{p}}_{vc}=\boldsymbol{p}_{vc}^{\frac{1}{\tau}}/\sum_{c=0}^{C-1}{\boldsymbol{p}_{vc}^{\frac{1}{\tau}}},
    \boldsymbol{\tilde{p}}_i=sharpen\left( \boldsymbol{p}_i,\tau \right) ={\boldsymbol{p}_{i}^{c}}^{\frac{1}{\tau}}/\sum_{c=0}^{C-1}{{\boldsymbol{p}_{i}^{c}}^{\frac{1}{\tau}}},
\end{equation}
where $\boldsymbol{p}_i^c$ represents the probability that node $i$ belongs to class $c$, $\tau$ is a hyper-parameter named temperature. The low temperature stimulates the model to produce lower-entropy predictions. When $\tau\rightarrow0$, $\boldsymbol{\tilde{p}}_{i}$ will approach a one-hot distribution. Then we evaluate the gap between the predicted probability distribution of all nodes and their corresponding anchors, and hope to reduce this gap during training, which is defined as a Low-Entropy loss for all nodes:

\begin{equation}
    LE\left( \mathbf{\tilde{P}},\mathbf{P} \right) =\sum_{i=0}^{N-1}{\left\| \boldsymbol{\tilde{p}}_{i}-\boldsymbol{p}_{i} \right\| ^2},
\label{eq:LE}
\end{equation}
where $\mathbf{\tilde{P}}=\{\boldsymbol{\tilde{p}}_0, \boldsymbol{\tilde{p}}_1, \dots, \boldsymbol{\tilde{p}}_{N-1}\}\in \mathbb{R}^{N\times C}$ and $\mathbf{P}=\{\boldsymbol{p}_0, \boldsymbol{p}_1, \dots, \boldsymbol{p}_{N-1}\}\in \mathbb{R}^{N\times C}$ are the anchor label and predicted probability distribution of all nodes respectively.

\textbf{Combination Loss:} In each epoch, we employ both the Cross-Entropy loss in Eq.\ref{eq:CE} and the Low-Entropy loss in Eq.\ref{eq:LE} as the final combination loss. Considering the randomness of the sparse strategy and the additivity of entropy, we can perform multiple random sparse on the generated soft-weighted adjacency matrix in each epoch, which increases the diversity to message passing. The loss function is expanded as follows:
\begin{equation}
    \mathcal{L} =\frac{1}{S}\sum_{s=1}^S{\left( CE\left( \mathbf{P}_{l}^{\left( s \right)},\mathbf{Y}_{l} \right) +\lambda LE\left( \mathbf{\tilde{P}},\mathbf{P}^{\left( s \right)} \right) \right)},
\end{equation}
where $S$ is the number of random graph sparse, $\lambda$ is a hyper-parameter controlling the balance of Cross-Entropy loss and Low-Entropy loss, and the anchor label $\mathbf{\tilde{P}}$ is updated as $sharpen\left( \frac{1}{S}\sum_{s=1}^S{\mathbf{P}^{\left( s \right)}},\tau \right)$.

\section{Experiments}
In this section, we conduct experiments with the aim of answering the following research questions.

\begin{itemize}[left=3em]
    \item [\textit{RQ1:}] Does our proposed DPGNN outperform the state-of-the-art GNNs? 
    \item [\textit{RQ2:}] Is the design of each part indispensable for DPGNN? 
    \item [\textit{RQ3:}] Does our improved message-passing paradigm show superiority over the traditional message-passing paradigm?
    \item [\textit{RQ4:}] Does differential control of node-specific message aggregation under different step lengths make sense?
    \item [\textit{RQ5:}] Do the settings of hyper-parameters impact the performance of DPGNN?
    \item [\textit{RQ6:}] Does our model expend a lot of training time?
\end{itemize}

In what follows, we foremost present the details of the experiment, including datasets, baselines, and implementation details. Then, we exhibit experiment results to reply the above six research questions.

\subsection{Experiments Setup}
\subsubsection{Evaluation Datasets} We evaluate our proposed model on six real-world datasets (Citeseer \cite{kipf2017semi}, UAI2010 \cite{wang2018unified}, ACM \cite{wang2019heterogeneous}, BlogCatalog \cite{meng2019co}, Flickr \cite{meng2019co}, and CoraFull \cite{bojchevski2017deep}) with different topological structures.
\begin{itemize}
    \item Citeseer is a public research citation network, with nodes representing papers and edges representing citation links. Attributes of each node are bag-of-words representations of the relevant paper.
    \item UAI2010 contains 3067 nodes in 19 classes and it has been tested in GCN for community detection.
    \item ACM is extracted from the original ACM database, where the nodes represent papers and the edges indicate that there are co-authors between two papers. Node features are bag-of-words of paper keywords.
    \item BlogCatalog is a blogger's social network, which is collected from the BlogCatalog website, and the node features are constructed by the keywords of user profiles.
    \item Flickr is an interest social network, where nodes represent users and edges represent relationships between users.
    \item CoraFull is the larger version of well-known Cora dataset, in which nodes are papers and edges represent the citation between node pairs.
\end{itemize}
The statistics of the six datasets are shown in Table \ref{tab:datasets}, from which we can see that Flickr and BlogDatalog are denser than the other datasets in topological structure.  The same conclusion can be directly observed from Figure \ref{fig:graph_structure}, from which we can observe the six graph datasets have different topological structures.

\begin{table}[tb]
\centering
\scalebox{0.9}{
\begin{tabular}{lrrrrrr}
\toprule
Datasets    & Nodes & Edges  & Classes & Avg Degree & Features \\ \midrule
Citeseer    & 3327  & 4732   & 6       & 2.84           & 3703     \\
UAI2010     & 3067  & 28311  & 19      & 18.46           & 4973     \\
ACM         & 3025  & 13128  & 3       & 8.68           & 1870  \\
BlogCatalog & 5196  & 171743 & 6       & 66.11          & 8189     \\ 
Flickr      & 7575  & 239738 & 9       & 63.30          & 12047    \\
CoraFull    & 19793  & 65311  & 70     & 6.60           & 8710     \\ \midrule
\end{tabular}}
\caption{The statistics of six datasets. The average degree is computed by $\frac{2M}N$.}
\label{tab:datasets}
\end{table}

\begin{figure}[tb]
\centering
\includegraphics[width=0.75\linewidth]{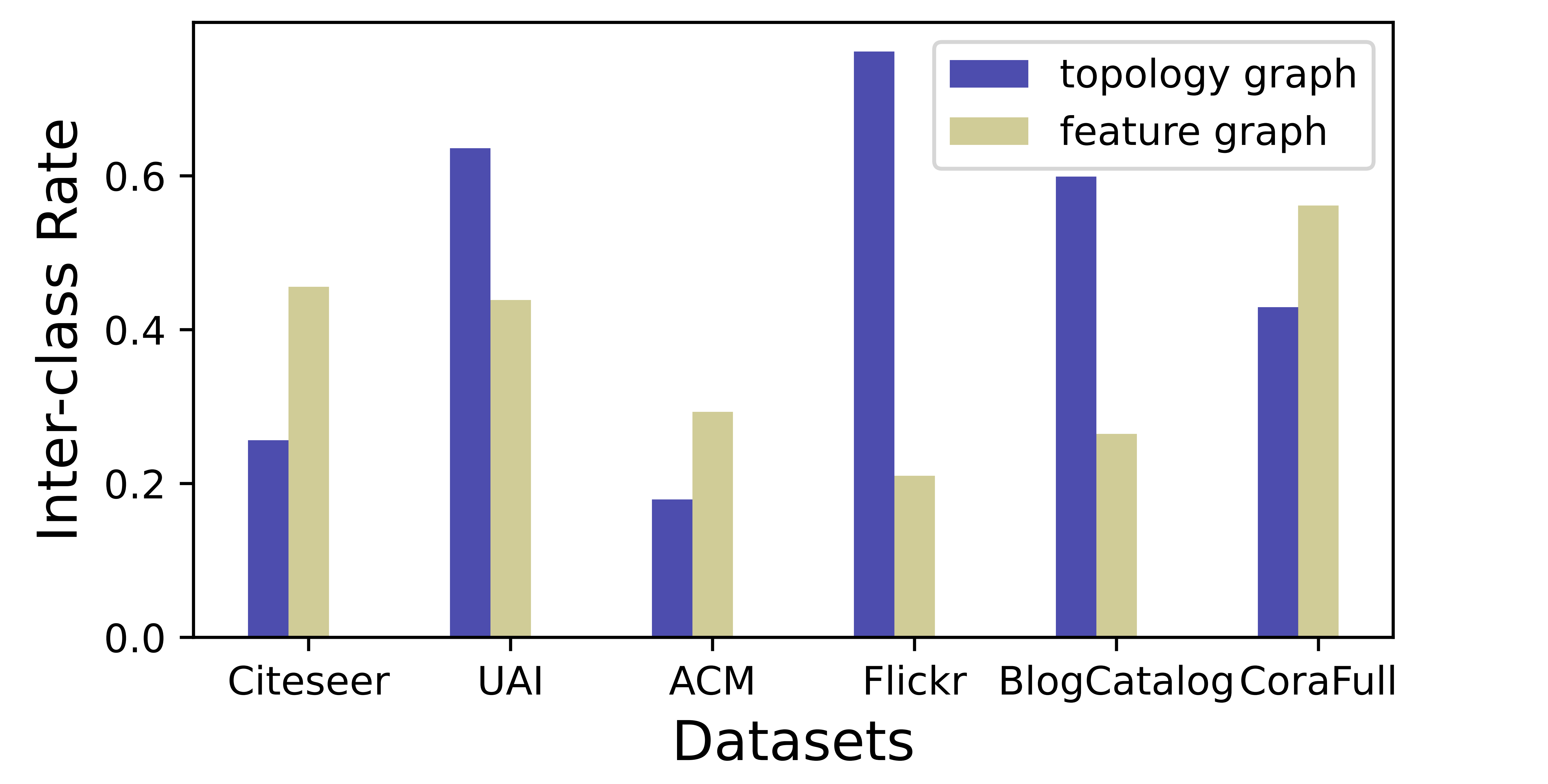} 
\caption{Inter-class rate comparison in topology graph and feature graph of six graph datasets.}
\label{fig:datasets}
\end{figure}

\begin{table*}[!ht]
\centering 
\scalebox{0.7}{
\begin{tabular}{lcccccc}
\toprule
Dataset   & \multicolumn{2}{c}{\textbf{Citeseer}}                & \multicolumn{2}{c}{\textbf{UAI}}              & \multicolumn{2}{c}{\textbf{ACM}}                         \\ \midrule
Metrics   & ACC   & F1    & ACC & F1  & ACC   & F1            \\ \midrule
MLP     & 60.6 ± 0.4  & 58.3 ± 0.4  & 65.9 ± 1.0  & 50.2 ± 1.5  & 76.3 ± 0.9  & 76.3 ± 0.8          \\ [2pt]
GCN     & 71.6 ± 0.3    & 68.2 ± 0.3    & 63.1 ± 0.7    & 50.0 ± 1.3    & 85.4 ± 1.3    & 85.6 ± 1.2          \\ [2pt]
SGC      &  71.9 ± 0.1 &  67.8 ± 0.3&63.9 ± 0.1 &  54.8 ± 0.1 &   83.2 ± 0.4&  83.4 ± 0.4          \\ [2pt]
APPNP   & 72.4 ± 0.5 & 68.6 ± 0.6    & 68.6 ± 0.9  & 53.5 ± 1.6  & 88.4 ± 0.7    & 88.4 ± 0.7          \\ [2pt]
DAGNN   & 73.3 ± 0.6  & 68.6 ± 0.5    & 64.9 ± 1.0    & 49.0 ± 1.3    & 88.2 ± 0.7    & 88.2 ± 0.7          \\ [2pt]
GCNII   & 73.4 ± 0.6  & 69.3 ± 0.6    & 64.2 ± 1.9    & 49.3 ± 2.9    & 89.1 ± 0.5    & 89.1 ± 0.5          \\ [2pt]
GRAND   & \uline{75.4 ± 0.4}   & \uline{70.2 ± 0.3}    & 67.9 ± 0.5    & 54.7 ± 0.7    & 88.0 ± 0.6    & 88.1 ± 0.5          \\ [2pt] 
AMGCN   & 73.1   & 68.4  & 70.1  & 55.6  & 90.4  & 90.4                \\ [2pt] 
SCRL   & 73.6    & 69.8  & \textbf{72.9}  & \uline{57.8}  & \uline{91.8}  & \uline{91.8}                \\ [2pt]
\textbf{DPGNN (Ours)}   & \textbf{76.2 ± 0.2}   & \textbf{70.4 ± 0.3} & \uline{71.1 ± 0.8} & \textbf{58.4 ± 1.0} & \textbf{92.5 ± 0.2} & \textbf{92.5 ± 0.2} \\[2pt] \midrule

\toprule
Dataset   & \multicolumn{2}{c}{\textbf{BlogCatalog}}         & \multicolumn{2}{c}{\textbf{Flickr}}    & \multicolumn{2}{c}{\textbf{CoraFull}}                         \\ \midrule
Metrics   & ACC   & F1    & ACC & F1  & ACC   & F1            \\ \midrule
MLP     & 74.0 ± 0.9  & 72.8 ± 0.8  & 54.3 ± 0.5  & 55.0 ± 0.5  & 46.9 ± 0.5  & 42.4 ± 0.5          \\ [2pt]
GCN     & 75.7 ± 0.4    & 74.5 ± 0.5    & 50.7 ± 0.5    & 50.1 ± 0.6    & 60.8 ± 0.4    & 55.9 ± 0.5          \\ [2pt]
SGC   &71.3 ± 0.1      &  70.1 ± 0.1 &  43.0 ± 0.1& 41.4 ± 0.1 &   57.4 ± 0.1  &    51.8 ± 0.1        \\ [2pt]
APPNP   & 82.0 ± 1.0 & 80.8 ± 1.1    & 58.3 ± 0.8    & 57.5 ± 0.9    & 57.2 ± 0.5    & 52.5 ± 0.5          \\ [2pt]
DAGNN   & 82.3 ± 3.4  & 81.3 ± 3.7    & 62.8 ± 1.4    & 62.4 ± 1.6    & \uline{60.5 ± 0.6}    & \uline{55.4 ± 0.8}          \\ [2pt]
GCNII   & 69.0 ± 4.1  & 69.1 ± 4.0    & 52.9 ± 1.5    & 55.3 ± 1.4    & 58.1 ± 0.5    & 53.1 ± 0.5         \\ [2pt]
GRAND   & 88.8 ± 0.9   & 87.9 ± 1.0    & 68.3 ± 0.5    & 67.5 ± 0.5   & 60.0 ± 0.3    & 53.0 ± 0.4          \\ [2pt]
AMGCN   & 82.0   & 81.4  & 75.3  & 74.6  & 58.9  & 54.7                \\ [2pt]
SCRL   & \uline{90.2}    & \uline{89.9}  & \uline{79.5}  & \uline{78.9}  & -     & -                \\ [2pt]
\textbf{DPGNN (Ours)}   & \textbf{90.9 ± 0.3}   & \textbf{90.4 ± 0.3} & \textbf{81.8 ± 0.4} & \textbf{81.9 ± 0.4} &  \textbf{62.7 ± 0.3} & \textbf{57.7 ± 0.6}  \\[2pt] \midrule

\end{tabular}}
\caption{Results on six datasets in terms of node classification accuracy (in percent) and F1 score. (Bold: best; Underline: runner-up)}
\label{tab:overall_resuts}
\end{table*}

Based on the previous analysis, the unfavorable message-passing structure is a disaster event for information aggregation. The high proportion of inter-class edges implies an imperfect message-passing structure. Given a graph structure $\mathcal{G} =\left( \mathcal{V} ,\mathcal{E} \right)$ of an arbitrary message space, the inter-class rate of edges is defined as follow:
\begin{equation}
\mathrm{ICR}_{\mathcal{G}}=\frac{\left| \left\{ \left( i,j \right) |\left( i,j \right) \in \mathcal{E} \,\,\land \boldsymbol{y}_i\ne \boldsymbol{y}_j \right\} \right|}{\left| \mathcal{E} \right|},
\label{eq:ICR}
\end{equation}
where $\boldsymbol{y}_i$ and $\boldsymbol{y}_j$ are the labels of node $i$ and $j$.
To further understand the original topology graph $\mathcal{G} ^t$ and the constructed feature graph $\mathcal{G} ^f$ of each dataset, we count their $\mathrm{ICR}_{\mathcal{G}^t}$ and $\mathrm{ICR}_{\mathcal{G}^f}$. The statistical results are shown in Figure \ref{fig:datasets}. All datasets have different inter-class rate in both spaces, which implies the diversity of experimental datasets. For example, the inter-class rate of Citeseer in topology space is lower than that in feature space. ACM has low inter-class rates in topology and feature space. Flickr has a extremely high inter-class rate in topology graph, while it has a quiet low inter-class rate in feature graph.

\subsubsection{Baseline} We compare DPGNN with eight methods, including a classic neural network: i.e., Multilayer Perceptron(MLP), two topology-based shallow GNNs, i.e. GCN \cite{kipf2017semi}, SGC\cite{wu2019simplifying}, four topology-based deep GNNs, i.e. APPNP \cite{klicpera2018predict}, DAGNN \cite{liu2020towards}, GCNII \cite{chen2020simple}, GRAND \cite{feng2020graph}, and two feature-based GNNs, i.e. AMGCN \cite{wang2020gcn}, SCRL \cite{liu2021self}.

\subsubsection{Implementation Details} Like most semi-supervised graph node classification tasks, we select 20 labeled nodes per class for training, 500 nodes for validation and 1000 nodes for testing. Specifically, in order for the impartial experimental comparison, we split the datasets are split in the same standard way as most topology-based GNNs, and make the data splits of UAI2010, ACM, BlogCatalog, Flickr equal to the feature-based model AMGCN. We conducted 100 runs with different random weight initialization for our proposed DPGNN, the results of which are averaged with $90\%$ confidence level. We perform a grid search \cite{Tang_UltraOpt} to tune our associated hyper-parameters for each dataset and apply the same standard for reproducing unreported datasets on other methods. We run all experiments on the Pytorch platform with Intel(R) Xeon(R) Gold 6140 CPU and GeForce RTX 2080 Ti GPU. We use Adam optimizer \cite{kingma2014adam} with learning rate 0.01 and weight decay 5e-4, and apply early stopping strategy. Specifically, we range step length $L$ from 2 to 8 for soft-weighted adjacency matrix learning in both topology space and feature space, and set $S \in \{1, 2,3,4,5,6\}$ for the random sparse operations.

\subsection{Performance Comparison (RQ1)}

The results on six datasets are summarized in Table \ref{tab:overall_resuts}, where our DPGNN generally achieves the best performance on all datasets compared with all the latest state-of-the-art models. Notably, all performance results have a small standard deviation, demonstrating the weak dependence on initialization parameters and the superior stability of our model. For all datasets, whether the topological structure is favorable or not, our model can achieve a competitive performance. Because our model can capture structural neighborhood information and feature-related information simultaneously, which proves the versatility of our proposed model. Particularly, for the Flickr dataset with poor topological structure, our model improves accuracy by at least $2.9\%$ and F1 score by at least $3.8\%$ compared with other methods. And we discover that most topology-based GNNs perform poorly on Flickr. The main reason is that Flickr's topological structure will produce a lot of noise in information propagation. Furthermore, most latest state-of-the-art models except GCNII perform well on Blogcatalog whose topological structure is slightly better than Flickr. We suspect that's because APPNP, DAGNN and GRAND decouple the two operations of representation transformation and information propagation, which alleviates the poor performance of topology structure to a certain extent. However, SGC, which also decouples these two operations, performs poorly on Flickr, even worse than MLP. That is because there is only a linear classifier for representation learning in SGC, and Flickr has a particularly high initial feature dimension.

\begin{table}[htb]
\centering
\scalebox{0.9}{
\begin{tabular}{lll}
\toprule
Model    & ACC   \\ \midrule
\textbf{DPGNN}    & \textbf{76.2 ± 0.2}  \\
\quad -w/o (1) parameter $\mathbf{W}_o$  & 75.1 ± 0.3 ($\downarrow1.4\% $)    \\
\quad -w/o (2) feature space  & 75.3 ± 0.5 ($\downarrow1.2\% $)      \\ 
\quad -w/o (2) topology space     & 71.4 ± 0.4 ($\downarrow6.3\% $)      \\
\quad -w/o (3) multiple random sparse   & 75.4 ± 0.3 ($\downarrow1.0\% $)     \\
\quad -w/o (4) random sparse strategy \& $S=1$   & 73.4 ± 4.2 ($\downarrow3.7\% $)   \\
\quad -w/o (5) Low-Entropy loss \& $S=1$   & 73.2 ± 0.4 ($\downarrow3.9\% $)\\
\\\midrule
\end{tabular}}
\caption{abalation study on dataset citeseer.}
\label{tab:abalation}
\end{table}

\subsection{Ablation Study for DPGNN (RQ2)}
To prove the contribution of each component of DPGNN, we conducted an ablation study on the public dataset Citeseer. We used the average accuracy as the standard. The result is shown in Table \ref{tab:abalation}.  We find that: (1) when we adopt the average attention instead of using the learning parameter $\mathbf{W}_o$ without considering the node-specific message output discrepancy, the accuracy drops $1.4\%$; (2) when we do not use the soft-weighted adjacency matrix $\overline{\mathbf{M}}^f$ based on feature space or $\overline{\mathbf{M}}^t$ based on topology space for message passing and representation learning, the accuracy drops $1.2\%$ and $6.3\%$ respectively, (3) when we only perform the random graph sparse once (i.e. $S=1$), the accuracy drops $1.2\%$; (4) when we do not apply random sparse strategy in the condition of $S=1$ (i.e. directly adopt $\mathbf{M}^t$ and $\mathbf{M}^f$), the accuracy drops $3.7\%$;  (5) when we do not introduce Low-Entropy loss in the condition of $S=1$ (i.e. directly set $\lambda=0$), the accuracy drops $3.9\%$. Above the aforementioned observation, each component of DPGNN contributes to the model, including classification accuracy and model stability. Particularly, case (1) illustrates the importance of focusing on the node-specific message output discrepancy. Case (2) shows that multi-space message interactions is complementary to graph representation learning, and our proposed message-passing paradigm can also achieve an ideal  performance on Citeseer without introducing the original topology space. Case (3) proves that the diversity of message passing contributes to the improvement of graph representation learning ability during training. Based on the comparison between case (3) and case (4), we find that the random graph sparse strategy plays a significant role in model stability. And case (5) demonstrates the significance of introducing Low-Entropy loss.

\begin{figure*}[htb]
\centering
\includegraphics[width=1.0\linewidth]{figure/ablation_2_space.png} 
\caption{The accuracy results of DPGNN compared with other specially designed models on six datasets.}
\label{fig:Property}
\end{figure*}

\subsection{Superiority of The Improved Message-Passing Paradigm (RQ3)}

In this part, we compare DPGNN with four specially designed models on all datasets to analyse the effects of the new message-passing paradigm in DPGNN.
\begin{itemize}
\item \textbf{DPGNN\_topo \& DPGNN\_fea}: DPGNN only based on topology space and DPGNN only based on feature space.
\item \textbf{GCN\_topo \& GCN\_fea}: GCN based on topology space and GCN based on feature space.
\end{itemize}

From the results in Figure \ref{fig:Property}, we can draw the following conclusions: (1) The results of DPGNN are generally better than other four models with single perception, indicating the validity of multi-space message interaction in the new message-passing paradigm. This part corresponds to the dual-perception representation learning design of DPGNN. (2) The performance of DPGNN based single-perception network is superior to that of GCN based single-perception network (i.e. DPGNN\_topo \textit{vs} GCN\_topo, DPGNN\_fea \textit{vs} GCN\_fea), verifying the availability of multi-step message source in the new message-passing paradigm. This part is related to the soft-weighted adjacency matrix learning design of DPGNN. (3) Citeseer, UAI2010 and ACM can benefit from both message-passing spaces simultaneously, BlogCatalog and CoraFull benefit mainly from message passing in topology space, and Flickr benefits mainly from message passing in topology space. This shows the diversity of the evaluation baseline datasets, and our proposed DPGNN can achieve competitive performance in both favorable and unfavorable graph topology datasets, demonstrating the universality of our proposed model. To prove the effectiveness of our proposed model more intuitively, we draw the final node representations on ACM dataset in Eq.\ref{eq:final_representation} by using t-SNE \cite{van2008visualizing}. Figure \ref{fig:DPGNN_tsne} shows the visualization of DPGNN performs best, in which the clearest distinct boundaries structures among different classes are exhibited compared to other methods.

\begin{figure*}[htb]
\centering
\includegraphics[width=1.0\linewidth]{figure/acm_tsne.png} 
\caption{The t-SNE visualization of node representations derived by DPGNN and four specially designed models on ACM dataset.}
\label{fig:DPGNN_tsne}
\end{figure*}

\begin{figure*}[tb]
\centering
\includegraphics[width=1\linewidth]{figure/control_of_output.png} 
\caption{The visualization of the output control coefficient ($o_{v\iota}$ in Eq.\ref{eq:soft-weighted}) learned in topology space and feature space for Citeseer dataset.}
\label{fig:n2h_att}
\end{figure*}

\subsection{Effects of node-specific message aggregation (RQ4)}
To confirm the significance of node-specific message output design, we visualize the output control coefficient ($o_{v\iota}$ in Eq.\ref{eq:soft-weighted}) learned in topology space and feature space for Citeseer dataset. The step-length $L^t$ in topology space  and the step-length $L^f$ in feature space for Citeseer are both set to 2. We selected 100 nodes to visualize. From Figure \ref{fig:n2h_att}, we can find that each node pays different attention to different-step message passing in topology space and feature space. Besides, the overall trend indicates that the attention value of the short-step messages is higher than that of the long-step messages. The aforementioned analysis implies that the differential control of node-specific message aggregation under different step lengths is  meaningful.

\subsection{Hyper-parameter Sensitivity (RQ5)}

In order to meticulously grasp the impact of hyper-parameters, we conduct experiments to investigate their influence. We firstly explore the influence of the count $S$ of random graph sparse during training on model performance. Then because the step length of paths affects the breadth of message source, we conduct experiments to explore the comprehensive impact of step length $L$ in both topology space and feature space. Lastly, we evaluate three aggregators for aggregating node representation generated in two message-passing spaces. Moreover, Table \ref{tab:Hyperparameters} reports the best hyperparameters of DPGNN we used for the results
reported in Table \ref{tab:overall_resuts}.

\begin{table}[htb]
\centering
\scalebox{0.7}{
\begin{tabular}{lcccccc}
\toprule
Hyperparameter    & Citeseer &UAI2010& ACM  & BlogCatalog & Flickr & CoraFull \\ \midrule
random graph-sparse times S    & 5  & 2   & 5       & 6           & 4 &2     \\
Step length $L^t$ in topology space     & 2  & 3  & 2      & 3    & 2 & 3    \\
Step length $L^f$ in feature space         & 3  & 6  & 4       & 2   &2        & 3  \\
Aggregator & attention  & attention & attention       & mean-pooling   & attention  & attention   \\  \midrule
\end{tabular}}
\caption{Hyperparameters of DPGNN for six datasets.}
\label{tab:Hyperparameters}
\end{table}

% Because the step length of paths affects the breadth of message source, we conduct experiments to explore the comprehensive impact of step length $L$ in both topology space and feature space. And we also explore the influence of the count $S$ of random graph sparse during training on model performance. Moreover, 

\subsubsection{Impact of $S$} 
To further verify the influence of hyper-parameter $S$ for the random graph sparse operation, we evaluate our model by setting different $S$. The corresponding experimental results are shown in Figure \ref{fig:sample}. As the $S$ value increases, the message passing becomes more diverse. We can observe that on all datasets, the results when $S>1$ are generally better than those when $S=1$, which demonstrates the effectiveness of multiple random graph sparse operations. With the introduction of multiple random graph sparse operation, the density of the generated soft-weighted adjacency matrix is reduced and the diversity of message passing is increased, which enhances the message-passing paradigm robustness. Besides, we discover that for all data sets, a small $S$ value enables the model to reach an ideal accuracy.

\begin{figure*}[htp]
\centering
\includegraphics[width=1\linewidth]{figure/sample.png} 
\caption{Analysis of the influence of the count $S$ of random graph sparse during training on model performance.}
\label{fig:sample}
\end{figure*}

\begin{figure}[ht]
\centering
\includegraphics[width=0.65\linewidth]{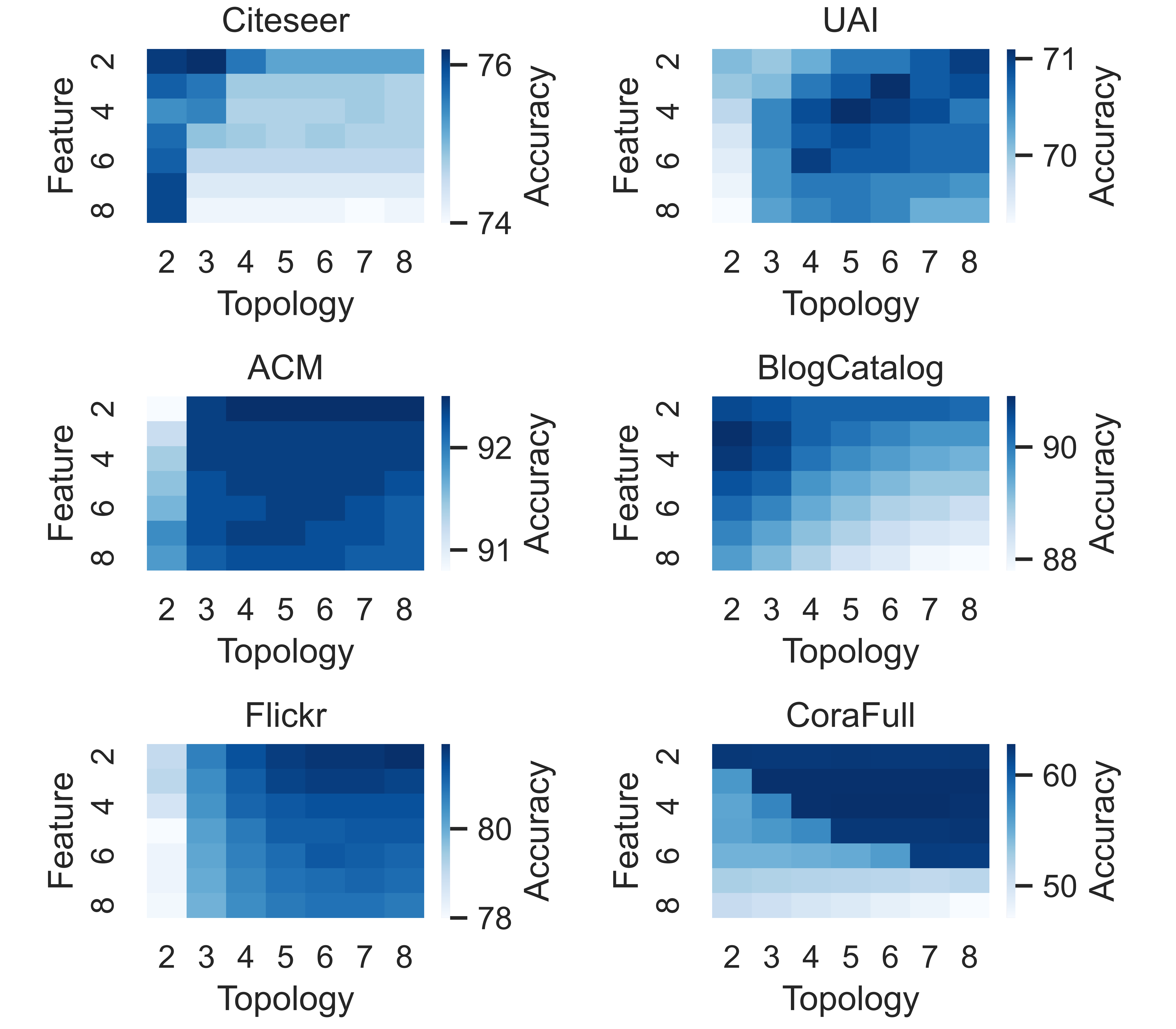}
\caption{Accuracy evaluation of the combinations of step length $L^t$ in topology space and $L^f$ in feature space. Both $L^t$ and $L^f$ are varied from 2 to 8.}
\label{fig:orders}
\end{figure}

\subsubsection{Impact of $L$} Step length settings in topology space and feature space determines the scope of message passing in the respective space, which can affect the performance of the model. Fig.\ref{fig:orders} shows the performance under different combinations of $L^t$ and $L^f$ on six datasets. We find that different datasets have different sensitivities to them. For example, Citeseer prefers $L^t$ set to 2, CoraFull prefers $L^f$ set to 2, and ACM tends to $L^t$ greater than 2.

\begin{figure}[ht]
\centering
\includegraphics[width=0.65\linewidth]{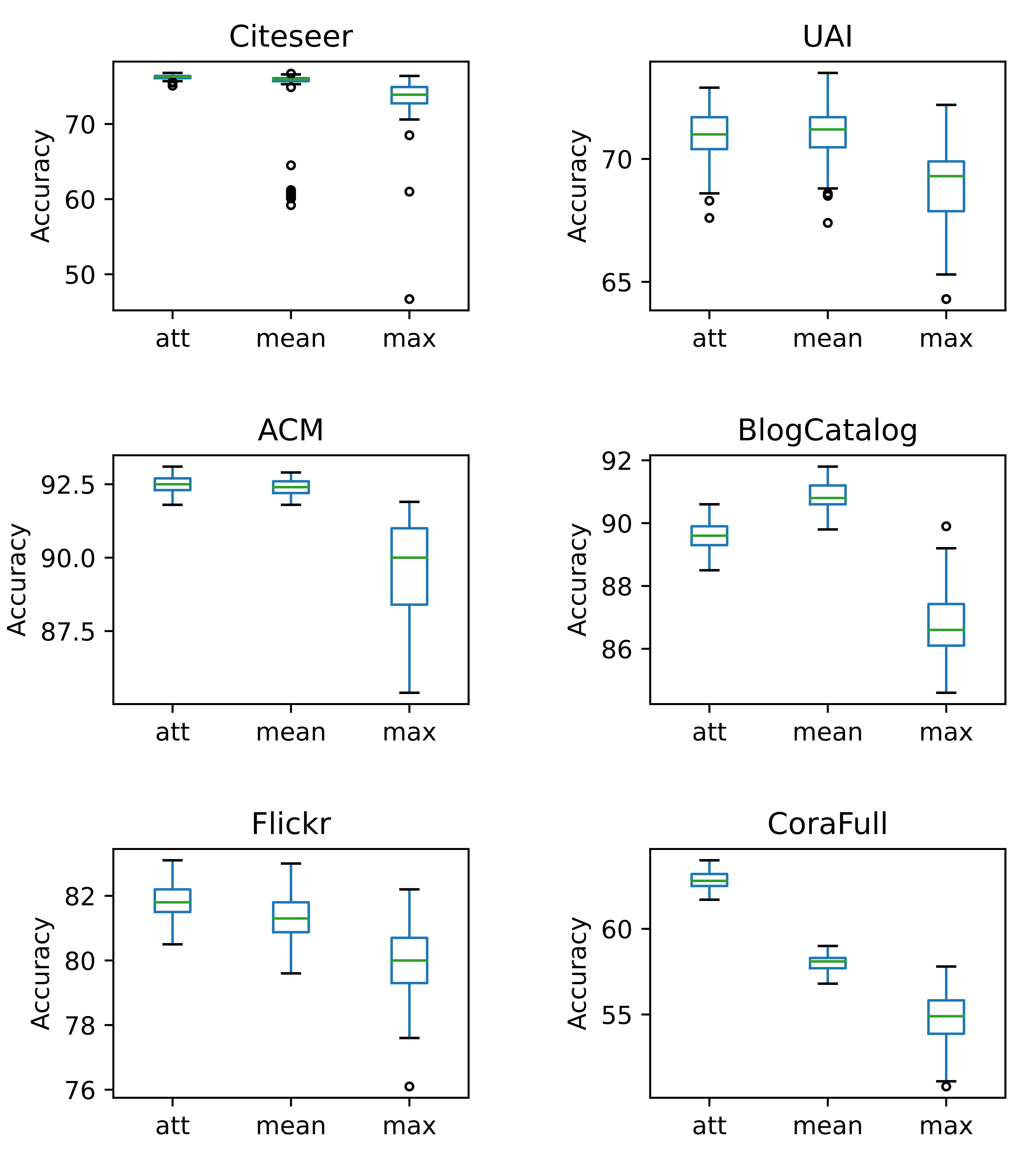}
\caption{Accuracy evaluation of three aggregators.}
\label{fig:agregator}
\end{figure}

\subsubsection{Impact of Aggregators} We evaluate three representation aggregators: Attention aggregator, Mean-pooling aggregator, and Max-pooling aggregator. As shown in Figure \ref{fig:agregator}, the Attention aggregator generally outperforms the other two aggregators except for BlogCatalog. The inferred reason is that Attention aggregator can aggregate node representations adaptively with its ability to distinguish the importance of node representation based on topology space and feature space. This also explains why our model can be adapted to datasets with different topological structures. Besides, Attention aggregator can also achieve a low small standard deviation, which means that Attention aggregator contributes to the stability of the model.

\begin{figure}[t]
\centering
\includegraphics[width=0.6\linewidth]{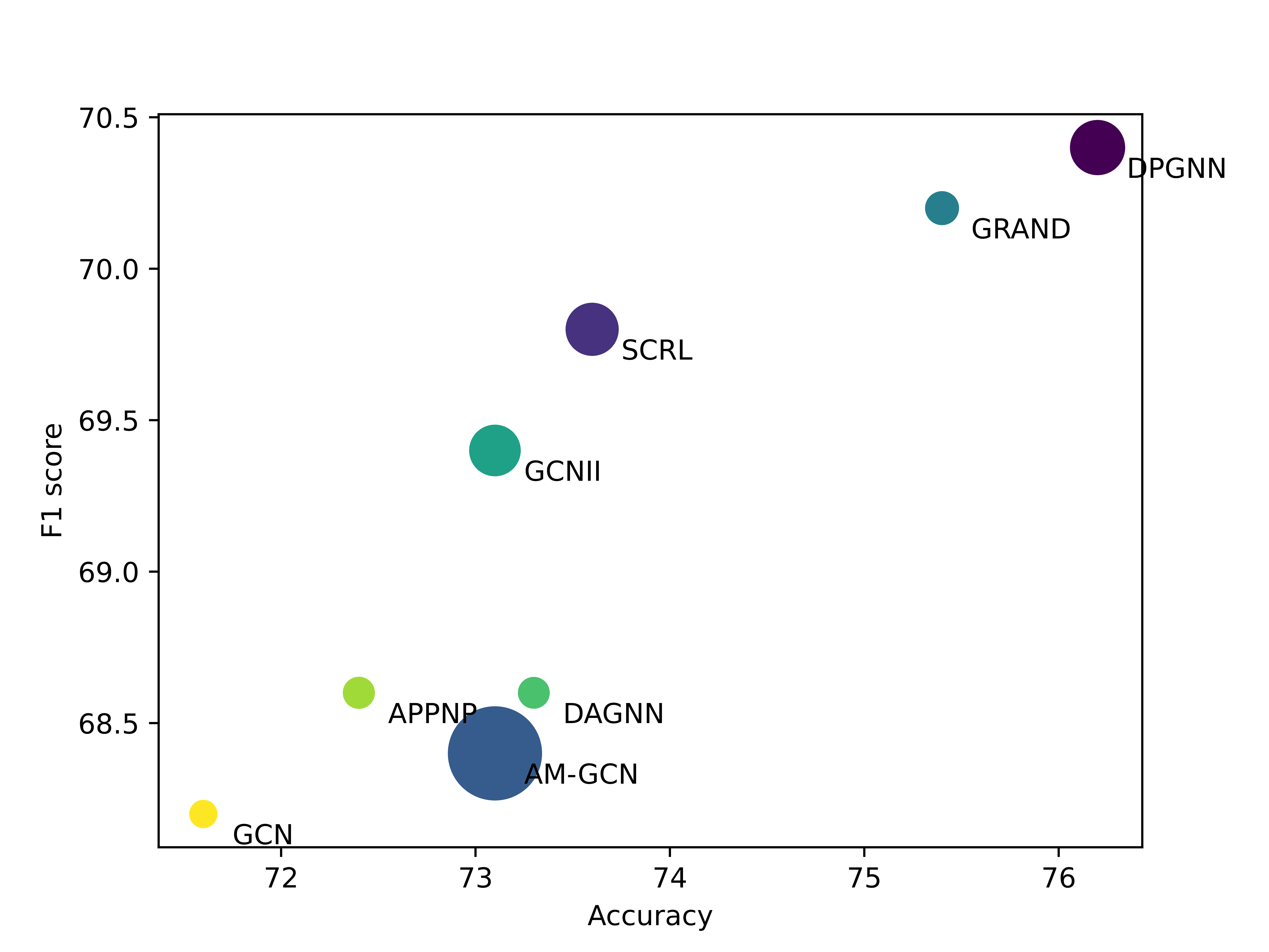}
\caption{A real running time comparison among several models.}
\label{fig:time}
\end{figure}

\subsection{Complexity Analysis (RQ6)} 
In our proposed DPGNN, the time complexity during training is mainly reflected in soft-weighted adjacency matrix learning and dual-perception representation learning. In the former part, the transfer coefficients can be precomputed, and because the 1×1 convolution operation can be parallelized across all nodes, the learning part can be computed with the time complexity $O\left( LN \right) $ , where $L$ is the step length and $N$ is the number of nodes. In the latter dual-perception representation learning part, the time complexity of the weight-share layer and two weight-exclusive layers is 
$O\left( Nd_h\left( d+2C \right) \right) $, and the time complexity of the message passing in topology space and feature space is $O\left( 2pLMC \right) $, where $d$ denotes the dimension of node feature, $d_h$ is the hidden size, $C$ is the number of classes, $M$ is the number of edges, and $p\in \left( 0,1 \right) $ is  the probability in random graph sparse strategy. From the aforementioned analyses, the time complexity of our DPGNN is linear with the number of nodes and edges. To realize the time complexity of our model more intuitively, we compare the real running time of several models on Citeseer dataset. We calculate the average running time of each training epoch for all models, which is shown in Figure \ref{fig:time}. The size of each point corresponds to the running time of each model. We observe that DPGNN can achieve competitive performance in an acceptable running time.

\section{Conclusion}
In this paper, we formalize the existing message-passing paradigm and analyze its drawbacks including inflexibility of message source expansion, negligence of node-level message output discrepancy, and restriction of single message space. To address these issues, we present a novel message-passing paradigm and verify it by instantiating a Dual-Perception Graph Neural Network (DPGNN). The main ingredients contributing to the success of DPGNN are: 1) broadening the multi-step message sources without extra iteration; 2) considering the node-specific message output discrepancy; 3) adopting the multi-space message interaction. We quantify the differences in topology and feature space of six graph datasets and conduct extensive experiments on them. The experimental results demonstrate that our instantiated DPGNN outperforms related state-of-the-art GNNs, and we conduct analysis experiments to prove the superiority and versatility of our proposed message-passing paradigm. To our knowledge, we are the first to consider node-specific message passing in the GNNs. In the future, we will explore more efficient methods for the node-specific message passing, and extend DPGNN to more challenging tasks.

\section{CRediT authorship contribution statement}
\textbf{Li Zhou}: Conceptualization, Methodology, Software, Formal analysis, Investigation, Data curation, Writing-original draft, Visualization. \textbf{Wenyu Chen}: Supervision, Investigation, Writing-Review \& Editing. \textbf{Dingyi Zeng}: Writing-Review \& Editing, Formal analysis. \textbf{Shaohuan Cheng}: Writing-Review \& Editing. \textbf{Wanlong Liu}: Visualization. \textbf{Malu Zhang}: Writing-Review \& Editing. \textbf{Hong Qu}: Supervision, Writing-Review \& Editing, Funding acquisition.

\section{Declaration of competing interet}
The authors declare that they have no known competing financial interests or personal relationships that could have appeared to influence the work reported in this paper.

\section{Acknowledge}
This work was supported by the National Science Foundation of China under Grant 61976043. 

% \section*{References}

\bibliography{mybibfile}

\appendix
\section{Instance Analysis}
\label{ap:1}
\begin{figure*}[htbp]
    \centering
    \subfigure[The definition of initial node feature]{
        \begin{minipage}[t]{0.3\linewidth}
            \centering
            \includegraphics[width=1.3in]{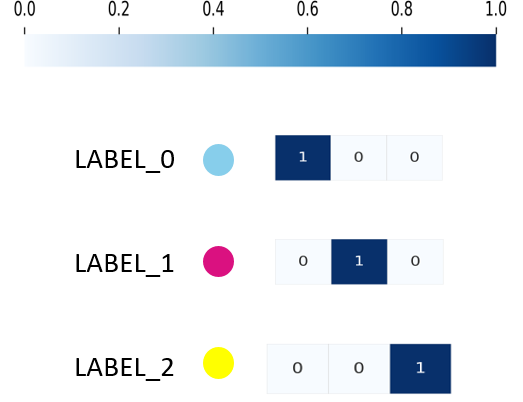}
            % \subsection*{11}
        \end{minipage}%
    }
    \subfigure[$\mathcal{G} _1\left( \mathcal{V} ,\mathcal{E} \right) $]{
    \begin{minipage}[t]{0.3\linewidth}
        \centering
        \includegraphics[width=1.3in]{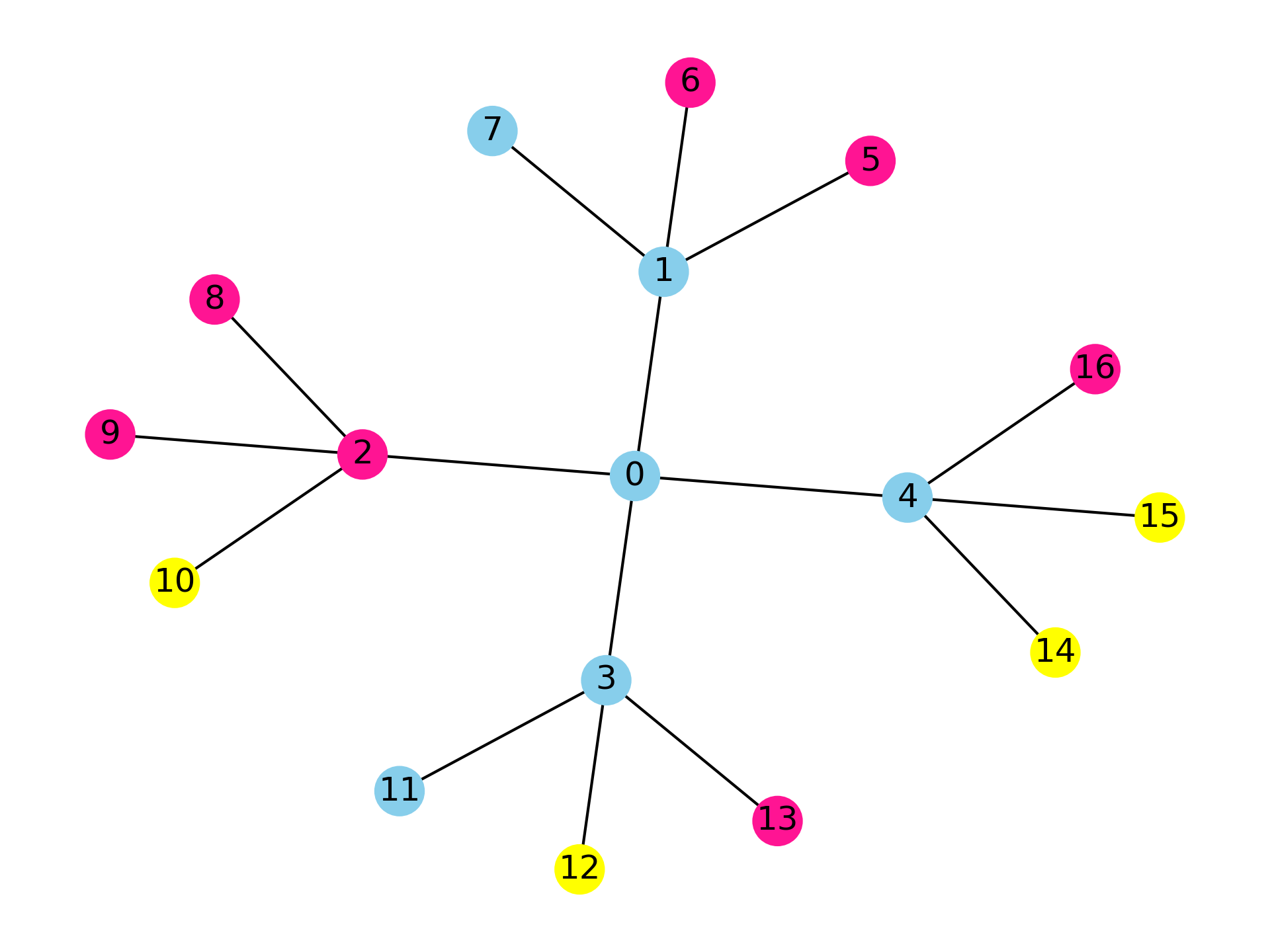}
        % \subsection*{12}
        \label{subfig:g1}
    \end{minipage}%
    }%
    \subfigure[$\mathcal{G} _2\left( \mathcal{U} ,\mathcal{E} \right) $]{
    \begin{minipage}[t]{0.3\linewidth}
        \centering
        \includegraphics[width=1.3in]{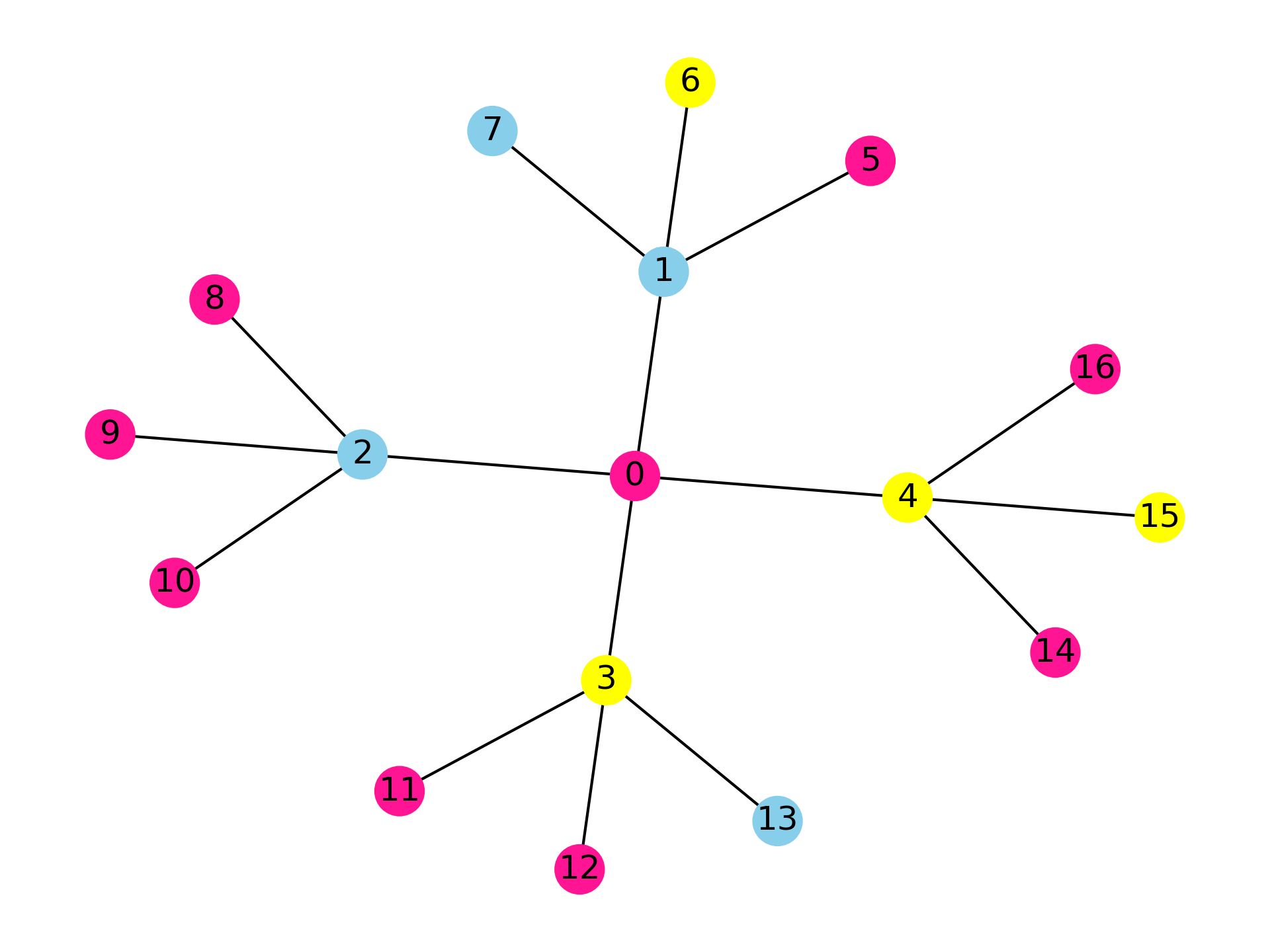}
        % \subsection*{13}
        \label{subfig:g2}
    \end{minipage}%
    }%
    \centering    
    \caption{Two 4-regular graphs and their node initial feature definition. The initial features of the central node $v_0$ in $\mathcal{G} _1$ and $u_0$ in $\mathcal{G} _2$ are represented as $\mathbf{m}_{v_0}^{\left( 0 \right)}=\left[ 1,0,0 \right]$ and $\mathbf{m}_{u_0}^{\left( 0 \right)}=\left[ 0,1,0 \right]$ respectively.} 
    \label{fig:two_graphs}
\end{figure*}

To clearly and intuitively verify that the existing message-passing paradigm ignores the amount of message output, we present two graphs with the same topology, denoted as $\mathcal{G} _1\left( \mathcal{V} ,\mathcal{E} \right) $ and $\mathcal{G} _2\left( \mathcal{U} ,\mathcal{E} \right) $ respectively, as shown in Figure \ref{fig:two_graphs}. $\mathcal{G} _1$ and $\mathcal{G} _2$ are mainly composed of 3 types of nodes. The nodes of the same label are drawn by the same color. We assume that the initial features of nodes with the same type are identical, and the initial features are defined into a 3-dimensional one-hot vector. 
By comparing Figure \ref{subfig:g1} and Figure \ref{subfig:g2}, we find that the central nodes $v_0 $ in $\mathcal{G} _1$ and $u_0 $ in $\mathcal{G} _2$ have the same neighborhood topology structure, but different neighborhood information distribution spaces. For $v_0$, its 1-hop neighborhood information is more important, while $u_0$ should concentrate more on its 2-hop neighborhood information.

In this experiment, we mainly focus on the central nodes $v_0 $ and $u_0 $ of $\mathcal{G} _1$ and $\mathcal{G} _2$, and detect their updated feature under different message-passing paradigms \footnote{Different from GNNs, we do not consider feature learning and only focus on message passing.}. We use the message-passing method $\mathbf{m}_{i}^{t}=\sum_{j\in \mathcal{N} (i)\cup i}{\frac{1}{\sqrt{deg\left( i \right)}\,\,\sqrt{deg\left( j \right)}}\mathbf{m}_{j}^{\left( t-1 \right)}}$ defined in GCN to materialize the amount of message retention $\alpha _i$ and the amount of message input $\beta _j$ in the existing message-passing paradigm (Eq. 1).

\begin{table*}[ht]
    \centering
    \scalebox{0.7}{
    \begin{tabular}{lccc}
        \toprule
        \textit{Line} & \textit{Message-Passing Paradigm}    & \textit{The updated feature of $v_0$ in $\mathcal{G} _1$} & \textit{The updated feature of $u_0$ in $\mathcal{G} _2$}\\ \midrule
        (1) & $\mathbf{m}_{i}^{\left( 1 \right)}=\alpha _i\mathbf{m} _{i}^{\left( 0 \right)}+\sum_{j\in \mathcal{N}_1 \left( i \right)}{\beta _j}\mathbf{m} _{j}^{\left( 0 \right)}$
        &
        \begin{minipage}[b]{0.3\columnwidth}
		\centering
            % \raisebox{-.5\height}{\includegraphics[width=\linewidth]{figure/G1_1_0.png}}
		\includegraphics[width=1.2in]{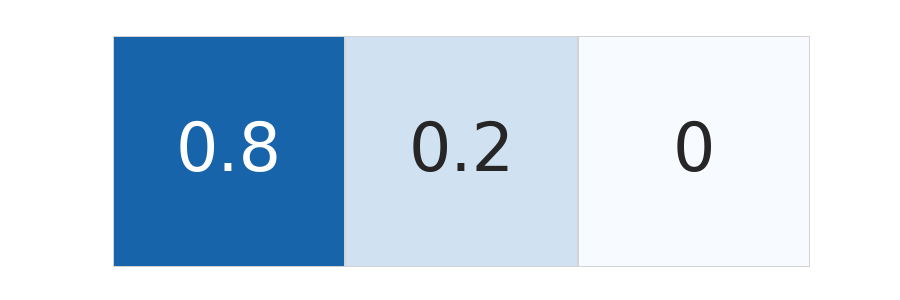}
	\end{minipage}
        &
        \begin{minipage}[b]{0.3\columnwidth}
		\centering
		\includegraphics[width=1.2in]{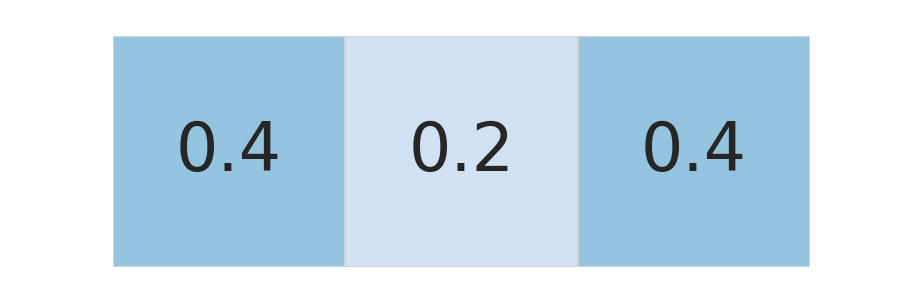}
	\end{minipage}  \\
        (2) & $\mathbf{m}_{i}^{\left( 2 \right)}=\alpha _i\mathbf{m} _{i}^{\left( 1 \right)}+\sum_{j\in \mathcal{N}_1 \left( i \right)}{\beta _j}\mathbf{m} _{j}^{\left( 1 \right)}$   
        &
        \begin{minipage}[b]{0.3\columnwidth}
		\centering
		\includegraphics[width=1.2in]{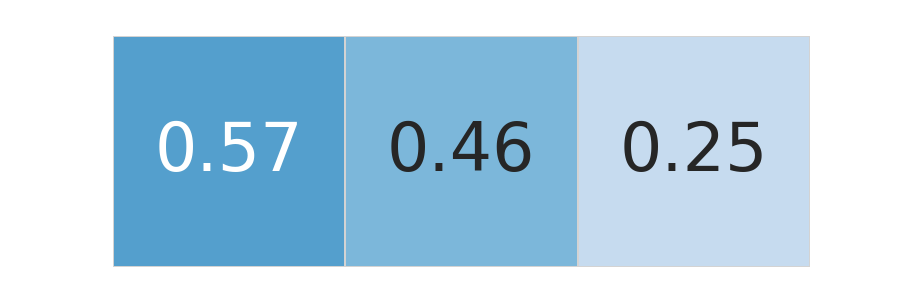}
	\end{minipage}
        &
        \begin{minipage}[b]{0.3\columnwidth}
		\centering
		\includegraphics[width=1.2in]{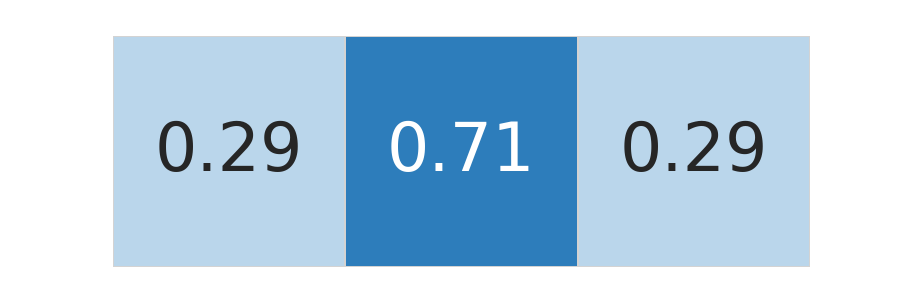}
	\end{minipage}  \\
        (3) & $\mathbf{m} _i=\sum_{\iota =0}^2{\left( \sum_{j\in \mathcal{J} _{\iota}\left( i \right)}{w_{\iota ,ij}}\mathbf{m} _j \right)}$    
        &
        \begin{minipage}[b]{0.3\columnwidth}
		\centering
		\includegraphics[width=1.2in]{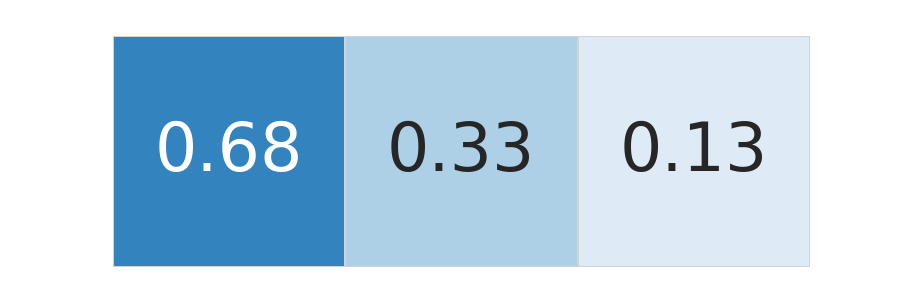}
	\end{minipage}
        &
        \begin{minipage}[b]{0.3\columnwidth}
		\centering
		\includegraphics[width=1.2in]{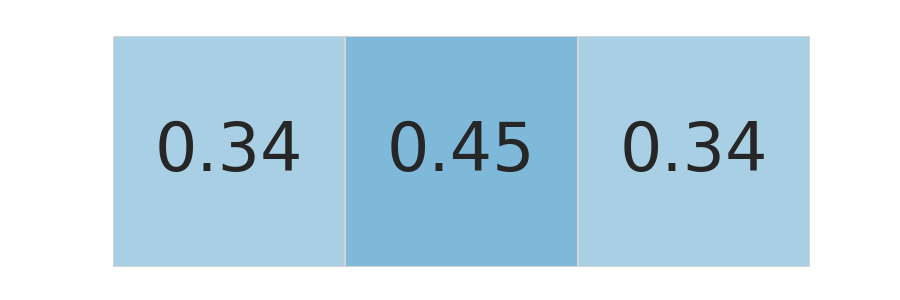}
	\end{minipage}  \\
         (4) & $\mathbf{m} _i=\sum_{\iota =0}^2{o_{i\iota}\left( \sum_{j\in \mathcal{J} _{\iota}\left( i \right)}{w_{\iota ,ij}}\mathbf{m} _j \right)}$    
        &
        \begin{minipage}[b]{0.3\columnwidth}
		\centering
		\includegraphics[width=1.2in]{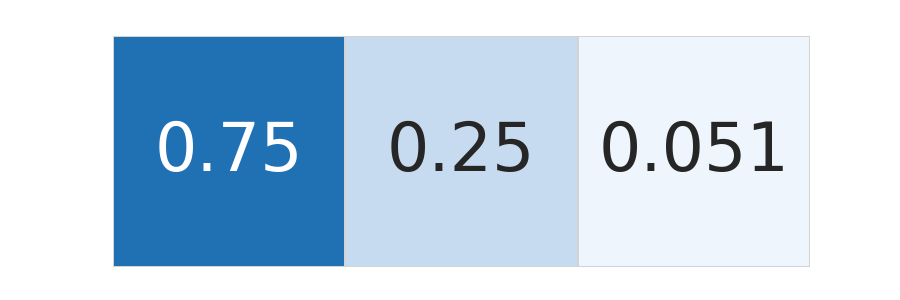}
	\end{minipage}
        &
        \begin{minipage}[b]{0.3\columnwidth}
		\centering
		\includegraphics[width=1.2in]{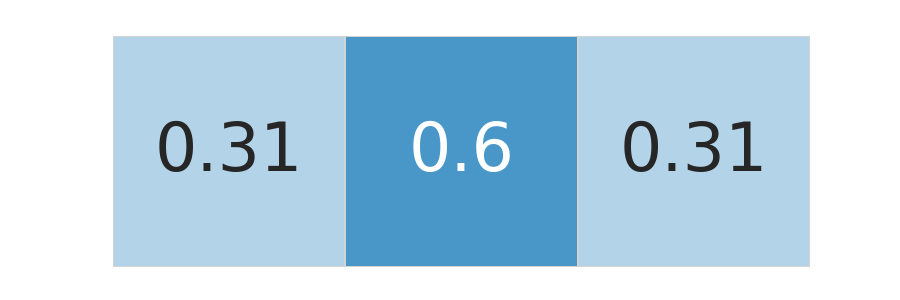}
	\end{minipage}  \\
 
    \bottomrule
    \end{tabular}}
    \caption{Visualize the updated features of $v_0$ and $u_0$ under different message-passing paradigms.}
    \label{tab:Visualize}
\end{table*}

As shown in Table \ref{tab:Visualize}, we visualize the updated features of $v_0$ and $u_0$ under different message-passing paradigms. In \textit{Line} (1), we follow the existing message-passing paradigm, and get the updated feature of $v_0$ and $u_0$ after one iteration, which shows $u_0$ abtain an undesirable feature because its 1-hop neighbors brings noise. We further show the results after two iterations in \textit{Line} (2). $u_0$ regains a desirable feature, while $v_0$ gains a non-ideal one. Therefore, the different neighborhood information space of each node is diverse, and the existing message-passing paradigm ignores node-level message output discrepancy, which is easier to bring noise information. To more intuitively demonstrate the advantages of our improved message-passing paradigm, we also visualize the related updated features. In \textit{Line} (3), the message-passing paradigm focuses on multi-step message source, and in \textit{Line} (4), our message-passing paradigm further considers the node-level message output discrepancy.

\
\end{document}